# Deep Learning-based Sentiment Analysis of Olympics Tweets

[1]Indranil Bandyopadhyay and [2]Rahul Karmakar

[1,2]Department of Computer Science, The University of Burdwan, Bardhaman, India

[1]banerjeeindranil143@gmail.com, [2]rkarmakar@cs.buruniv.ac.in

**Abstract:** *Sentiment analysis (SA), is an approach of natural language processing (NLP) for determining a text's emotional tone by analyzing subjective information such as views, feelings, and attitudes toward specific topics, products, services, events, or experiences. This study attempts to develop an advanced deep learning (DL) model for SA to understand global audience emotions through tweets in the context of the Olympic Games. The findings represent global attitudes around the Olympics and contribute to advancing the SA models. We have used NLP for tweet pre-processing and sophisticated DL models for arguing with SA, this research enhances the reliability and accuracy of sentiment classification. The study focuses on data selection, preprocessing, visualization, feature extraction, and model building, featuring a baseline Naïve Bayes (NB) model and three advanced DL models: Convolutional Neural Network (CNN), Bidirectional Long Short-Term Memory (BiLSTM), and Bidirectional Encoder Representations from Transformers (BERT). The results of the experiments show that the BERT model can efficiently classify sentiments related to the Olympics, achieving the highest accuracy of 99.23%.*

**Keywords:** Sentiment Analysis, Natural Language Processing, Neural Network, Deep Learning, Naïve Bayes, Convolutional Neural Network, Bidirectional Long Short-Term Memory, Bidirectional Encoder Representations from Transformers, Olympics.

## 1. INTRODUCTION:

The Olympic Games, or the Olympics (French: *Jeux olympiques*) [1], a pinnacle of athletic excellence and international camaraderie, testify to the sport's capacity to unite people globally. These respected events attract competitors and fans worldwide, breaking down geographical, cultural, and linguistic barriers. The advance digital age has brought a new era of involvement in the Olympics, with social media platforms acting as lively hubs of contact and conversations. There is a global upsurge of colloquialisms, dialogues, slangs, and viewpoints as social media sites like Twitter, Instagram, Facebook, and others come alive with debates, critiques, and sentimental statements about various aspects of the Olympic Games. Among these platforms, Twitter, now known as 𝕏 is a popular place where people can freely express their thoughts and emotions, creating a vibrant online community of Olympic lovers [2]. In 2024 Twitter/ 𝕏 has over 368 million monthly active users worldwide [3].

The Olympics take place every 4 years. According to the International Olympic Committee (IOC) reports, thousands of athletes compete in various games throughout the Winter and Summer Games. These competitions aim to promote global goodwill and highlight the limits of human potential rather than just medal competitions [4].

The study basically aims to understand the mood of the global community about the Olympics. The Olympics have always been a trending topic on social media, where fans express their excitement, support, and sometimes disappointment. We also analyze recent tweets to gauge the global mood for the upcoming Paris games, considering organizational changes.

The research does not endorse any particular nation or athlete but rather is based on social media discourse.

Iconic moments that influenced sport's history most of occurred in the Olympics, including in Berlin 1936 Jesse Owens' 4 gold medals [5], Perfect 10 by Nadia Comăneci in gymnastics in 1976 [6], Usain Bolt's lightning thunderbolt in Beijing 2008 and London 2012 [7], and the historic gold medal win of Neeraj Chopra in javelin throw at the Tokyo 2020 Olympics, making him the first Indian athlete to win an Olympic gold [8], etc. Whether it's the thrill of a world record being broken, or the collective pride in national achievements, or the rise of a new athletic hero, the Olympics continue to be a powerful force that unites people from around the world. However, without overindulging in Olympic domain our main focus is on the advance DL models on how we can achieve the highest accuracy for SA.
Previous research highlights the need to improve how we analyze social media discussions about global events. There is a lack of studies on Olympic-related tweets, and those that do often suffer from low accuracy, there are many challenges in using Twitter data for SA and predicting actual public interest. To address these gaps, our study combines cultural studies, and social sciences and uses NLP, DL, contextual embeddings, and robust feature extraction to improve accuracy and reliability.
As data and computing power have increased in the past few years, the field of artificial intelligence (AI) has seen a gradual increase in interest in DL, a branch of machine learning (ML) and it is based on artificial neural networks (ANN) that draw inspiration from the function and structure of the brain [9]. Models for DL that extract dynamics and features directly from the data can expedite the data preparation process and have a superior comprehension of intricate data patterns [10].

In essence, this paper aims to capture the global sentiment, reflected in Twitter data about the Olympics, using advanced DL models and shedding light on the complex aspects. We used a large dataset of tweets.

## 1.1 Background:

Before delving deeply into SA using DL, it is necessary to understand some basics.

- *Neural Networks:*

Neural Networks (NNs) consist of many neurons, which are layered information processing units collaborating to carry out tasks (such as classification) by modifying the weights of connections, similar to how a biological brain learns. Through training, these networks may uncover intricate links, patterns, and relationships in data [11]. (Fig-1) shows a NN's basic structure.

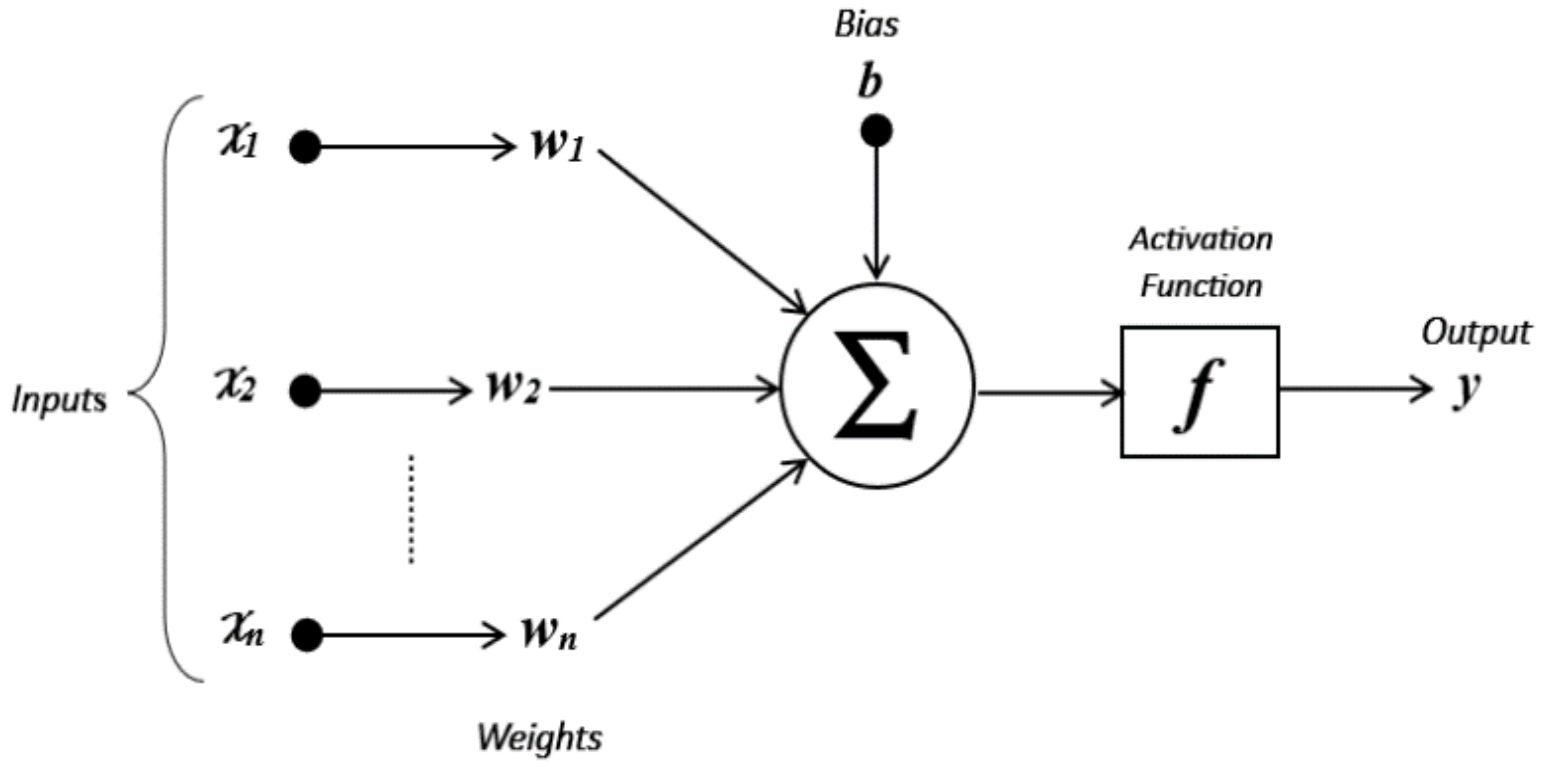

**Figure 1:** Basic Structure of a Neural Network

Neurons (the basic computing elements of NN), accept inputs, multiply them by weights (for transformation), and then generate outputs. They produce outputs by including the bias terms, summing the weighted input values, and applying the activation function. NNs consist of layers like the input, hidden, and output layers; weight defines connection strength. Biases enable neurons to change their output irrespective of inputs; they define the impact of input signals. The NN may learn more intricate patterns by introducing non-linearity, which is made possible via activation functions. Some of the most frequently used activation functions are tanh, sigmoid, ReLU (Rectified Linear Unit), and softmax [12].

- *Deep Learning:*

In Deep Learning (the subset of ML), the adjective "deep" indicates, in the network the use of multiple layers [13]. So, to simulate the complex decision-making power of the human brain, it uses a multilayered strategy that conforms to the hidden layers of the NN, called Deep Neural Networks (DNNs). Features are defined and extracted manually or by feature selection techniques in traditional ML approaches. DL models, on the other hand, achieve higher accuracy and performance since features are learned and extracted automatically. Generally, the hyperparameters of classifier models are also being measured automatically [14].

A comparison of sentiment polarity classification between traditional ML and DL is shown in (Fig- 2).

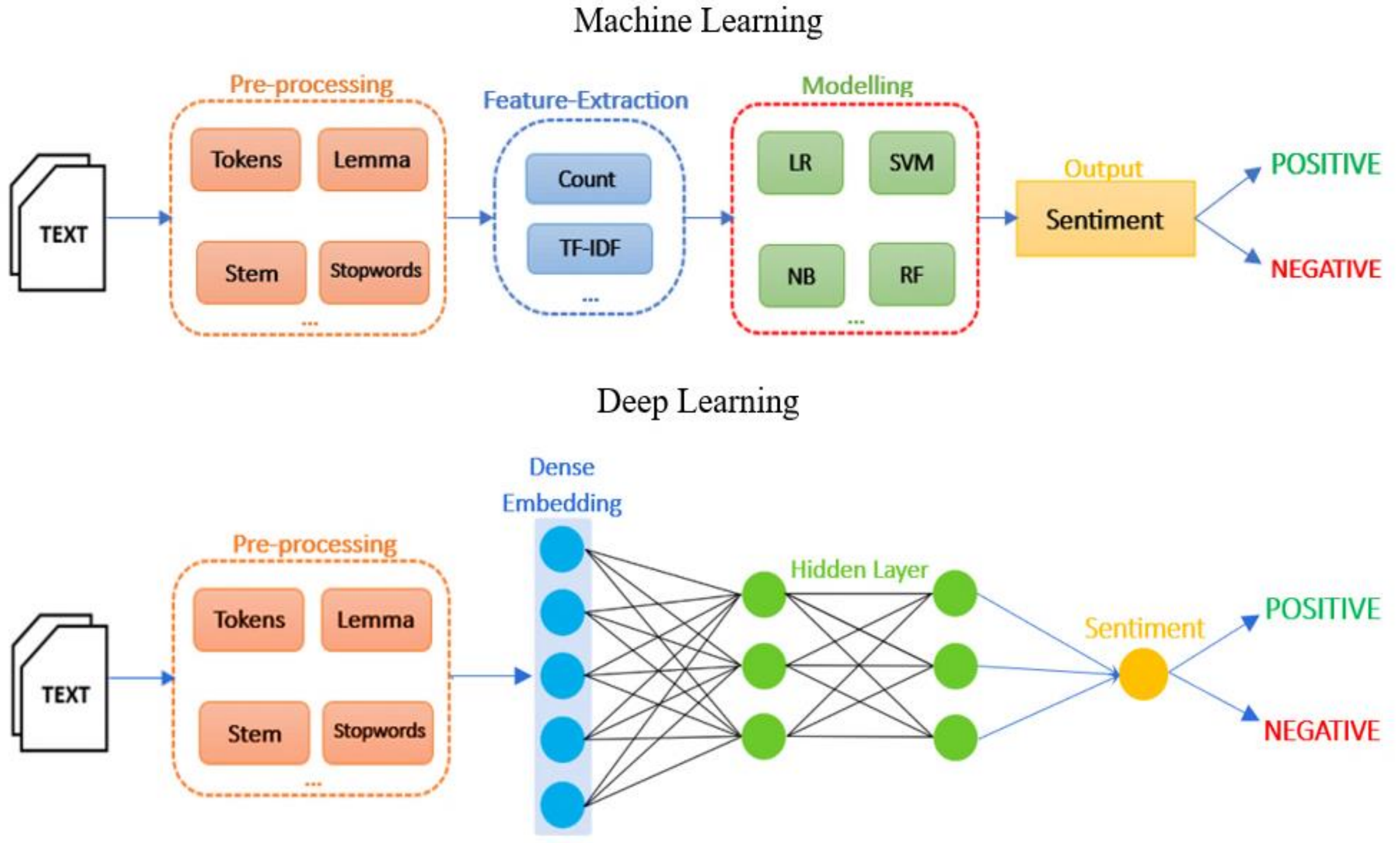

**Figure 2:** Two learning approaches for sentiment polarity, ML (top), and DL (bottom).

Traditional ML models are simpler and faster to train, suitable for limited computational resources or time, but due to manual feature engineering constraints, generally struggles with large datasets. But, on the other hand, DL models use numerous hidden layers to learn complex, hierarchical data representations. It is capable of fine-tuning pre-trained models for smaller datasets using transfer learning. DL facilitates end-to-end learning, requiring significant computational power. These models use advanced regularization techniques and dynamically adjust to different input data types. Traditional and deep learning differences include feature engineering, model complexity, computational requirements, training data needs, scalability and adaptability, and interpretation [15].

To many problems in the realm of image and speech recognition, as well as in NLP, NNs and DL models currently provide the best solutions. Some types of advance DL techniques are discussed in this study.

***"Deep learning is going to be able to do everything" – Geoffrey Hinton*** [16].

Each of CNNs, BiLSTMs, and BERT has special advantages in SA. CNNs are efficient and do feature extraction well especially on short texts, while BiLSTMs provide a robust understanding of sequential and long-range dependencies in the texts, and with the Transformer based architecture, BERT is a huge leap in NLP as it captures the contextual information and deep contextual understanding. Out of these models, BERT is shown to be the most accurate model because of the bidirectional context capture and fine-tuning capability, making it a powerful tool for analyzing sentiment in text.

### 1.2 Structure of the Paper:

The content of this study (document) is structured as follows, described in (Fig- 3). At first, The Introduction and Background in Section 1, and then we evaluated the topical literature review in Related Work Section 2. In Section 3 the proposed Methodology and Experiments, the suggested models are evaluated, and their effectiveness is in Section 4. In Section 5, the conclusion is presented along with a few suggestions for future work.

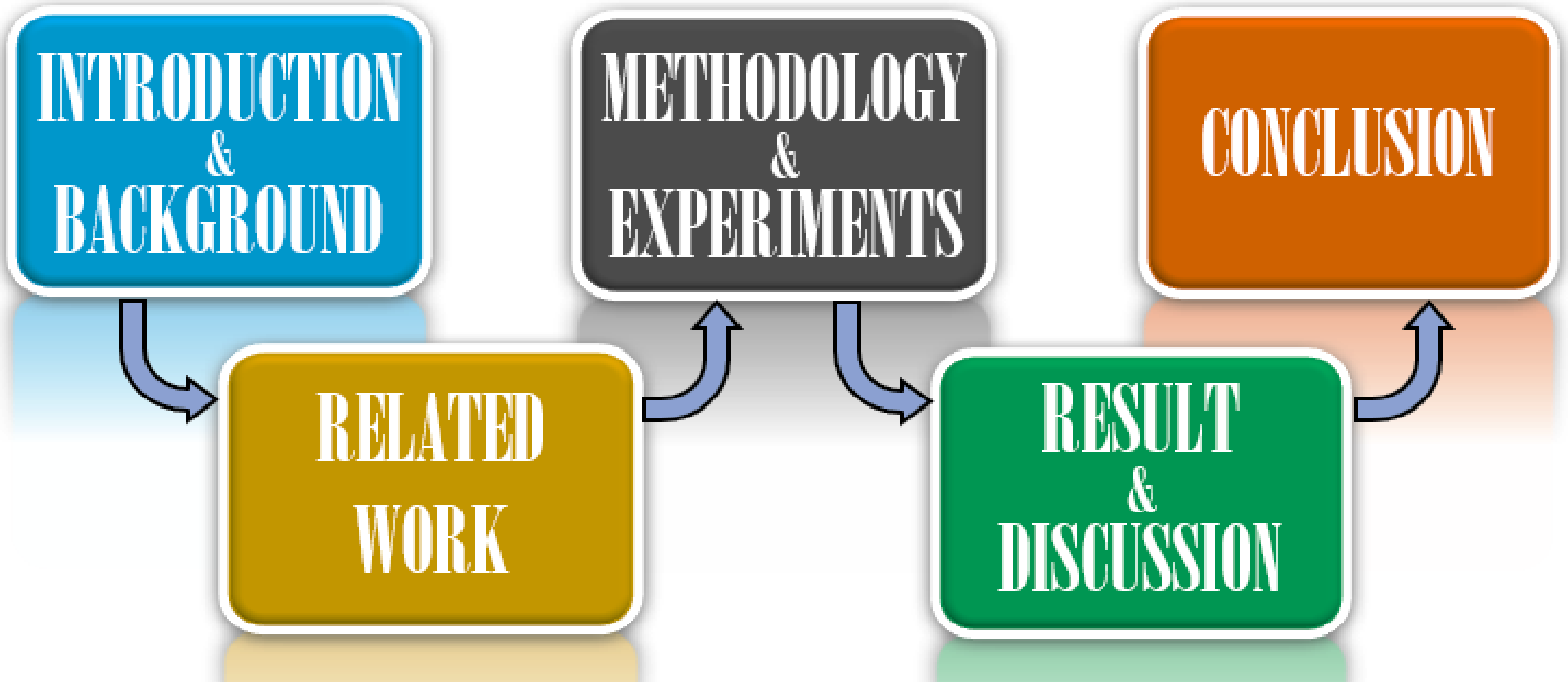

**Figure 3:** The Illustration of Paper Structure

## 2. RELATED WORK:

Several studies were undertaken to acquire expertise in evoking emotion through text. Before commencing this research, we focused on examining numerous significant varieties of studies to gather insights. Very little work has been done on sentiment analysis from Olympics Twitter data. Therefore, to gain knowledge, as SA has a vast application domain, let's go through the various related works that have been done.

Research has shown that SA using DL can reveal interesting insights into public opinion surrounding major sporting events (e.g., the Olympics). The SA results suggest the importance of DL techniques in assessing public sentiment and market reactions also.

This study [17] provides a comprehensive analysis of SA using DL-based models. The paper also compares the enhancements of DL architectures and their issues and factors for enhancing the accuracy of SA. The study identifies capsule-based RNN approaches with an accuracy of 98.02%. The innovative CRDC model demonstrated superior performance across different databases, for the Toxic dataset CRDC model gives the best accuracy of 98.28%.
In this study, [18] CNNs are used to categorize English movie reviews. Two FE techniques, the BoW model and TF-IDF are used to convert reviews into a numerical representation. The BoW model achieves 84.84% classification accuracy in the proposed CNN framework, while TF-IDF has better precision at 88.96%.
Domain-independent research has also been done, in this research paper [19] a domain-independent SA model is created by training the model on various domain-independent datasets. By addressing constraints such as feature extraction and dataset size; this work focuses on DL-based models for sentiment classification. Five models, including CNN-GRU, CNN-LSTM, CNN, LSTM, and GRU, are trained. Test data from three datasets and a new book review dataset were used to assess performance. The GRU model achieved the highest accuracy of 80.18%.
This study [20] analyzes tweets and the IndoBERT model to investigate the public's perception of the 2024 Indonesia general election. The IndoBERT model for SA achieves 83.5% accuracy.

Aspect-based Sentiment Analysis (ABSA) detects sentiment towards specific service, or experience aspects, using DL methods for real-time applications like customer feedback analysis and social media monitoring. This study [21] focuses on creating ABSA models using DL concepts in order to find sentiments regarding particular characteristics or components of a service, product, or experience. There are three DL models used: CNN, BiLSTM, and Hopfield Network (HN).
In this paper [22], a total of 10,000 tweets were collected and trained for the Tokyo 2020 Olympics public health SA using an LSTM model. The tweets had been classified as positive, negative, or neutral. The model's sentiment classification final validation accuracy was 88.2%. Positive sentiments about public health during the Tokyo 2020 Olympics were 40.39%.
In this research paper [23], Using SA, the authors explored how global public opinion has changed and how sentiment has varied during the four Olympic Games. The worldwide public opinion setting and epidemiological background of the Beijing Winter Olympics in 2022 are comparable to those of the Beijing Olympics in 2008, the Sochi Winter Olympics in 2014, and the 2020 Tokyo Olympics. User-generated content (UGC) from Chinese and English social media platforms between 2008 and 2022 was analyzed, and the results revealed a trend toward general politicization and an increase in the use of strong sentimental statements. Chinese individuals have more negative attitudes about politics than English people, although they have more positive attitudes toward athletic activities.
The study [24] uses #LancsBox, the SVM algorithm, and Python data mining to examine sentimental tendencies on Twitter-related topics. The findings indicate that Twitter users are more likely to hold negative views regarding the Beijing Winter Olympics.
This study [25] presents a method for using SA to interpret news articles inside a multilingual corpus. It evaluates and combines SA algorithms to improve output quality. The study reviews the limits of the algorithms using SHAP and proposes using the 3 classifiers (Amazon BERT, Sent140 BERT, and Vader) to identify contradictory results. An imbalance in the media's coverage of the legacies of London 2012 and the Rio 2016 Olympics is exposed by utilizing a dataset of 1271 articles in Portuguese and English from 7 media sites.
This Case study [26] proposes for destination image (DI) research a text-mining framework using UGC and SA methods. It focuses on the 2022 Winter Olympics and identifies 9 image attributes, including ticketing service and beginner suitability. The study finds a continuous downward trend in negative tourist sentiment over the past 7 snow seasons, while positive sentiment shows a slow upward trend. Tourists believe, Winter Olympics will improve their destination's image if it meets their expectations.
This study [27] investigates the flow of SA. It explores common ML and DL (LSTM, CNN) techniques. With word embedding method, CNN model got highest accuracy of 84.5% after a longer training time.
In this paper [28] the DL models classifies positive emotions and negative emotions. TF-IDF and Doc2Vect are used and got the highest accuracy of 91.3% with LSTM model for neg subclass.
This research [29] investigates the impact of word count and review readability on sentiment classification performance. The study found that high readability and brief reviews performed best using DL techniques like SRN, LSTM, and limiting text length was more effective for accuracy.
In this paper [30] inspired by MRC on physical documents, the authors developed an RRC dataset and developed a post-training technique on the BERT language model to improve performance.
This paper [31] uses 2077610 tweets in Persian and English and clustered using the Hive tool (in Hadoop). Results from the 2016 Olympic and Paralympic games demonstrate good recall and precision. Response times are reduced with big data processing technologies such as Pig and Hive.
Due to data features like tweet length, spelling mistakes, and special characters, SA on social media platforms is challenging task. This paper [32] proposes a CNN model that outperforms baseline models.
This article #London 2012 [33] suggests a combination method for analyzing major sporting events using social media data. It considers event days, user groups, SA, and topic extraction. The results reveal that diverse geographical and temporal patterns may be found, discriminating between residents and tourists and extracting issues about transportation infrastructure.

In this informative paper [34], Tang and Zhang discuss the success of NN approaches in SA, including sentiment-specific word embeddings, sentence and document classification, and fine-grained tasks.
In this research paper [35] a CNN model for emotion recognition in sports-related tweets has been developed. 3 significant games and the 2014 FIFA World Cup were covered by the model, which was used to examine Twitter tweets. It achieved 55.77% accuracy after being trained on automatically tagged tweets in seven distinct moods.

**Table 1:** Summary of Related Work

| *Author and years* | *Method* | *Dataset* | *Result or, (Accuracy)* | *Limitations* |
|---|---|---|---|---|
| *Shofiqul et al., 2024[17]* | Comparative analysis, CRDC model gives the highest accuracy | IMDB<br>Toxic<br>CrowdFlower<br>ER | 88.15%,<br>98.28%,<br>92.34%,<br>95.48%. | The limitations are difficulty in accurately discerning emotions, and handling semantics and linguistic nuances, and some basic limitations in traditional methodologies and deep learning techniques are there. Addressing these issues could lead to further advancements in SA using DL models. |
| *Lou, 2023 [18]* | BoW,<br>TFIDF, CNN model | English movie reviews | 84.84%,<br>83.48%. | The limitations are small dataset size, generalizability, feature extraction, hyperparameter tuning, interpretability, and evaluation metrics. Addressing these could improve the model's robustness, generalizability, and interpretability. |
| *Qamar et al., 2023 [19]* | CNN-GRU, CNN-LSTM, CNN, LSTM, and GRU models give the highest accuracy | IMDB movie review<br>Amazon book review | 84.32%<br>80.18% | The model's performance could be improved by finding the best hyperparameters for training and using different word embeddings. |
| *Geni, Evi and Indra, 2023 [20]* | IndoBERT model | Election Twitter data | 83.5% | 1k tweets are less and data imbalance is there as it's dominated by neutral tweets. |
| *Bharathi, Bhavani, Priya, 2023 [21]* | HN,<br>CNN,<br>BiLSTM | Benchmark dataset for ABSA | 75.69%,<br>92.08%,<br>89.46%. | The limitations are limited dataset description, lack of model comparison scope, overlooking evaluation metrics, generalizability, and lack of feature engineering techniques. |
| *Salau et al., 2023[22]* | LSTM model | Tokyo2020 Olympics public health tweets | 88.20% | The limitations are not exploring more sources, potentially overlooking diverse viewpoints and not utilizing more advanced models for sentiment analysis. |
| *Jia et al., 2022 [23]* | Topic modeling (LDA), TF-IDF, NB model | Olympics data collected from Twitter, YT, and news media | English sentiment is more positive (overall) | The limitations are, not completely capturing the opinions diversity in the general population and not mentioning accurately the model's performance. |
| *Hou, 2022 [24]* | Python data mining, #LancsBox, SVM model | Beijing Winter Olympics 2022 Tweets | 62.9% negative, 37.1% positive tweets | The limitations are limited dataset description, lack of model comparison, and not mentioning the model's performance. |
| *Mello, Gullal, Gaurish, 2022 [25]* | Amazon BERT | Olympic legacy news articles dataset. | 74.7% | The limitations are translation issues, difficulty in handling negations, inability to recognize sarcasm, and the tendency for news headlines to be more negative than the corresponding texts. |
| *Peng et al., 2022 [26]* | LDA model (based on custom lexicon) | UGC Olympic reviews through web crawler | destination image analysis insights | The limitations are very short review texts, only used Chinese data, and image perception was limited to relevant reviews. |
| *Umarani, Julian, Deepa, 2021 [27]* | NB,<br>LR,<br>CNN,<br>LSTM | Restaurant reviews | 76.42%,<br>73.6%,<br>84.5%,<br>77%. | The DL models (CNN and LSTM) outperformed the ML classifiers in terms of the accuracy but, the main drawback of their DL models is overfitting, and the generalization problem. |
| *Shilpa et al., 2021 [28]* | TF-IDF, Doc2Vec, LSTM,<br>RNN model | Tweets of pos/neg, pos subclass, neg subclass. | 91.3%,<br>87.02%. | The limitations are no extensive tweet cleaning, limited availability of dataset, polysemy issues, and noisy labels introduce inaccuracies in the training process and affect the overall performance. |
| *Li et al., 2020 [29]* | SRN, CNN,<br>LSTM | IMDb Movie Review dataset | Approx. 60%,<br>Over 70% | The limitations are focusing on a single dataset, limited consideration of other textual features, and exclusive use of deep learning models. |
| *Xu et al., 2019 [30]* | BERT-PT | ReviewRC from SemEval 2016 | 78.07% | The study may have limitations in terms of generalizability, small dataset size, low annotation quality, low computational resources, and task complexity. |
| *Seilsepour, 2019 [31]* | Big Data SA (Pig & Hive) | Olympic Games 2016 Tweets dataset | 78.99% (Precision) | The limitations are relying on individual words, uncertainty about competition participation, and difficulty in accurately interpreting sentiment signals. |
| *Alharbi, Doncker, 2019 [32]* | J48 (DT),<br>LSTM,<br>CNN | Twitter datasets published by SemEval-2016 | 87.61%,<br>88.13%,<br>88.71%. | The limitations of models and methodologies include unbalanced datasets, reliance on feature selection, complexity of DL models, and generalizability. |
| *Kovacs-Gyori et al., 2018 [33]* | Hu Liu lexicon, and LDA for topic modeling. | Geolocated tweets from London during the 2012 Olympic Games dataset | Clear sentiment classificati on and empirically derived LDA | The limitations include restricted data availability to 2012 London tweets, potential user location bias, limited language consideration, and the empirical derivation of LDA parameters may not capture all topic nuances. |
| *Tang and Zhang, 2018 [34]* | Discussions about various DL models such as CNN, RNN, LSTM, etc. | Discussions about common datasets like IMDb movie reviews, Twitter sentiment data, etc. | Successful NN approaches for SA overview. | The paper discussed various limitations of SA using DL models may include data bias, interpretability challenges, subjective data annotation, computational resource requirements, and domain adaptation issues but, other important issues are not mentioned in this research paper. |
| *Stojanovoski et al., 2015 [35]* | Word2Vec, GloVe. CNN model | The dataset is based on Wang et al.'s work. | 55.77% highest accuracy for 10,000 training samples | The limitations are using shallow CNN architecture, potential issues with word embeddings, dataset imbalance, and the scope of accurate generalization in emotion classification for tweets. |

# 3. METHODOLOGY & EXPERIMENTS:

We frame sentiment analysis using deep learning models as a supervised classification task. Our objective is to classify Olympics tweets, and that dataset was gathered from Kaggle (kaggle.com). Relevant features are extracted from the tweets for sentiment categorization, and deep learning methods are employed along with a baseline model, shown in (Fig- 4).

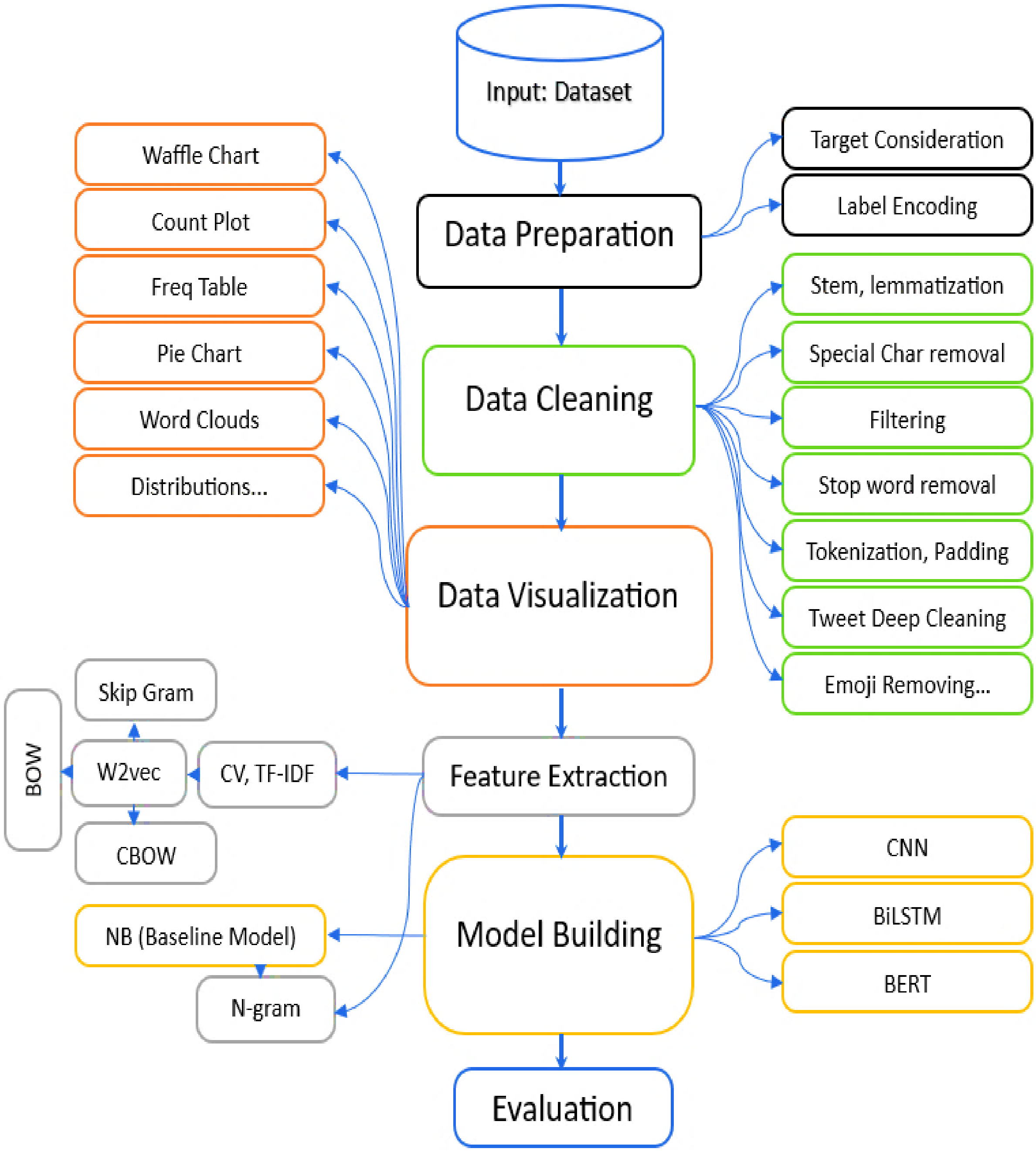

**Figure 4:** Proposed SA Methodology

- *The SA Workflow:*

The workflow of the framework for SA is illustrated in (Fig- 4), and it comprises the main seven phases of operation, i.e., Choosing a good dataset, data preparation, cleaning, visualizations, feature extraction, model building, and presentation of the result.

A brief description of its phases is given below.

### 3.1 Dataset Selection and Preparation:

It is an obvious fact that, in the realm of ML/DL, the acquisition of data is the initial and pivotal stage, with a profound understanding of datasets before the inception of ML/DL solutions. A dataset - the foundational input in any ML/DL endeavor, comprises various digitally stored information.

***"In God we trust, all others must bring the data" - W. Edwards Deming*** [36].

Data provides clarity and objectivity, so it results in better decisions. However, emotions and personal biases can cause tunnel vision, myopic perspectives, and costly mistakes. Raw data must be transformed into usable format by making it more digestible, requiring a certain level of data literacy.

***"Data beats emotions." - S. Rad*** [37].

- *Dataset Selection:*

For our research, we used a carefully selected dataset from Kaggle (kaggle.com), specifically tailored to encompass tweets related to the Olympics. The dataset selection procedure is important, as it has a significant impact on the outcomes of our DL models. We can focus on analysis and model-building, by using an existing dataset. The dataset contains Olympics tweets and we know Twitter/ 𝕏 is a rich source of user-generated content. We ensured that our dataset was suitably prepared and refined.

- *Dataset Preparation:*

One of the most critical and also time-consuming tasks is data preparation, requiring meticulous attention to detail. According to research by Whang et al. [38], a substantial number of data scientists and AI engineers invest approximately 70% of their time in data collection and preparation, emphasizing its critical significance in shaping the quality and performance of DL models.

***"Data is just like crude oil. It's valuable, but if unrefined it cannot really be used." - M. Palmer*** [39].

Data preparation is crucial for successful ML/DL projects as it directly influences model performance and accuracy. It involves collecting data from various sources, cleaning it, transforming it through normalization and encoding, and reducing complexity without losing information.
Data preparation is a continuous process that requires revisiting and refining steps as the model evolves or new data is acquired. Two common myths regarding data preparation are that it's a one-time task and that more data is always preferable! [40].
Data preparation in a manual way is considered the gold standard, but it can be time-consuming, and susceptible to human error, but automated tools can handle it more efficiently and quickly.

***"The goal is to turn data into information, and information into insight." - Carly Fiorina*** [41].

- *Sentiments (target) consideration:*

We separated the target variable and the features. In this case, the sentiment of the tweets was considered the target variable, while the tweet text served as the feature. For supervised learning models this step is very crucial where the target variable guides the training process.

***"What we want is a machine that can learn from experience." - Alan Turing*** [42].

- *Label Encoding:*

To ease DL, we used label encoding to convert the categorical target variable into a numerical representation. Specifically, we encoded positive sentiments as 1 and negative sentiments as 0. For learning models to effectively process and learn from the data, this numeric encoding is essential.

***"Numbers have an important story to tell,<br>They rely on you to give them a clear and convincing voice." - Stephen Few*** [43].

The CSV file contains two columns: text and sentiment and has 2,77,801 rows.

| | A | B |
|---|---|---|
| 1 | text | sentiment |
| 2 | rejected sports 2020 olympics include bowling chess tug war | 0 |
| 3 | team usa men basketball team playing horrible olympics going 2004 make far | 0 |
| 4 | my_hive_away im watching kayaking olympics broadcast cool sport enjoy watching | 1 |
| 5 | girls age still trying figuring open account onlyfans 18 years old anastasija zo | 1 |
| 6 | omensfromeden patty mills olympics | 0 |
| 7 | one fav mario party mini games olympics | 0 |
| 8 | leannewhittle1 derventioexcel jacobwhittle11 olympics teamgb incredible swim proud parent day | 1 |
| 9 | becauseimatter agree feels like focus elsewhere meaning excitement olympic | 0 |
| 10 | ms dhoni shooting gun olympics vachestundi | 0 |
| 11 | whenever watch victory ceremonies years olympics know official music composed | 0 |
| 12 | djokersa supersporttv olympics games different rights example argentina one state chann | 0 |
| 13 | still called 2020 olympics | 0 |
| 14 | watching olympics badminton look ping pong need drink | 0 |
| 15 | highlight evening pakistani weightlifter talha talib almost medal country | 0 |
| 16 | thinking ended 2nd heartbroken | 0 |
| 17 | merlisa amid chaos surrounding olympics naomiosaka poised create happy ending winning gold | 1 |
| 18 | redsqkk seantrende lack respect country us suck olympics best | 1 |
| 19 | olympics rigged | 0 |
| 20 | watching tiktok videos athletes olympics village kinda makes feel nostalgic id like | 0 |
| 21 | really wish nike wouldve waited olympics start releasing usa sb x parra kits something f | 1 |
| 22 | anyone watching olympics tv viewers drop 33year low daily wire | 0 |
| 23 | profound slap face corrupt world professional cycling | 0 |
| 24 | congrats dr kiesenhofer phd mathematics first win olympic medal austria | 1 |
| 25 | tokyo olympics roommates shushiladevi mirabai_chanu send emotional thank message manipur | 0 |

**Table 2:** Shows the CSV file (the dataset)

### 3.2 Data Loading and Cleaning:

- *Data loading:*

We have to load our data. In Google Colab, datasets can be easily loaded from sources like Google Drive using Python libraries like *pandas*. For the BiLSTM model, our 100% dataset is loaded at once. The dataset loading can also be done by loading the already separated train_dataset (90%) and test_dataset (10%), ensuring generalization to unseen data, used for our CNN and BERT model. Proper loading and splitting help to build robust DL models.

- *Data Cleaning:*

Effective data cleaning is necessary, this ensures the accuracy and reliability of DL models. This portion details the steps and methodologies employed in cleaning the dataset, transforming it from raw, noisy data into a structured, analyzable format. We took the following steps to clean the data:

First, to standardize the text data, we handled the contractions. A CSV file containing common contractions and their expanded forms was read into a DataFrame and converted into a dictionary for efficient replacement. For the sake of uniformity, all contractions and their meanings were converted to lowercase.

Several regex patterns were defined to identify and replace specific sorts of text elements, including URLs, @user mentions, #hashtags, sequences of characters, and emojis. The *emojis* are a common way for individuals to express their emotions, so we must pay special attention to them. These patterns allowed for the systematic identification and replacement of these elements.

To apply the cleaning steps to each tweet, *preprocess_apply*, a preprocessing function, is implemented. This function carried out multiple operations: it converted the tweet to lowercase, replaced URLs with the token *<url>*, replaced user mentions with the token *<user>*, replaced sequences of three or more identical characters with two characters, replaced *emojis* with descriptive tokens, expanded contractions using the predefined dictionary, removed non-alphanumeric characters and symbols, and ensured spaces were placed around slashes to separate words.

More data cleaning steps included:

1. We took abbreviations and acronyms from the text and replaced them with their full forms, e.g., “LOL” was expanded to “laugh out loud.”
2. Words containing digits were converted to text format, e.g., “N8” was converted to “night.”
3. Common stop words, which do not add significant value to the text sentiment, were removed.
4. The Lemmatization process reduced words in their root forms (lemma), e.g., “running” becomes “run”, enhancing the consistency of the text data.
5. A special phonetic algorithm was used to normalize words that had different spellings but the same pronunciation.

To meet specific cleaning needs, other custom functions were defined in addition to these steps. We removed unwanted *emojis* also using a function that strips unwanted *emoji* characters from the text.

Another function filtered out unwanted special chars, punctuations, links, mentions, and new line characters and to remove multiple spaces from the text a final function was used. Next, the preprocessing function was then applied to the entire dataset to clean each tweet methodically.
This comprehensive data-cleaning procedure significantly improved the text data's quality, reducing noise and improving the input's quality for our models.

- *Tweet Deep Cleaning:*

Moreover, tweet deep cleaning also has been done, we created a column to display the cleaned text length so that we could see if we removed too much text or almost entirely the tweet!
We have used the separately loaded datasets train_data and text_data for this visualization, mainly for clarification.

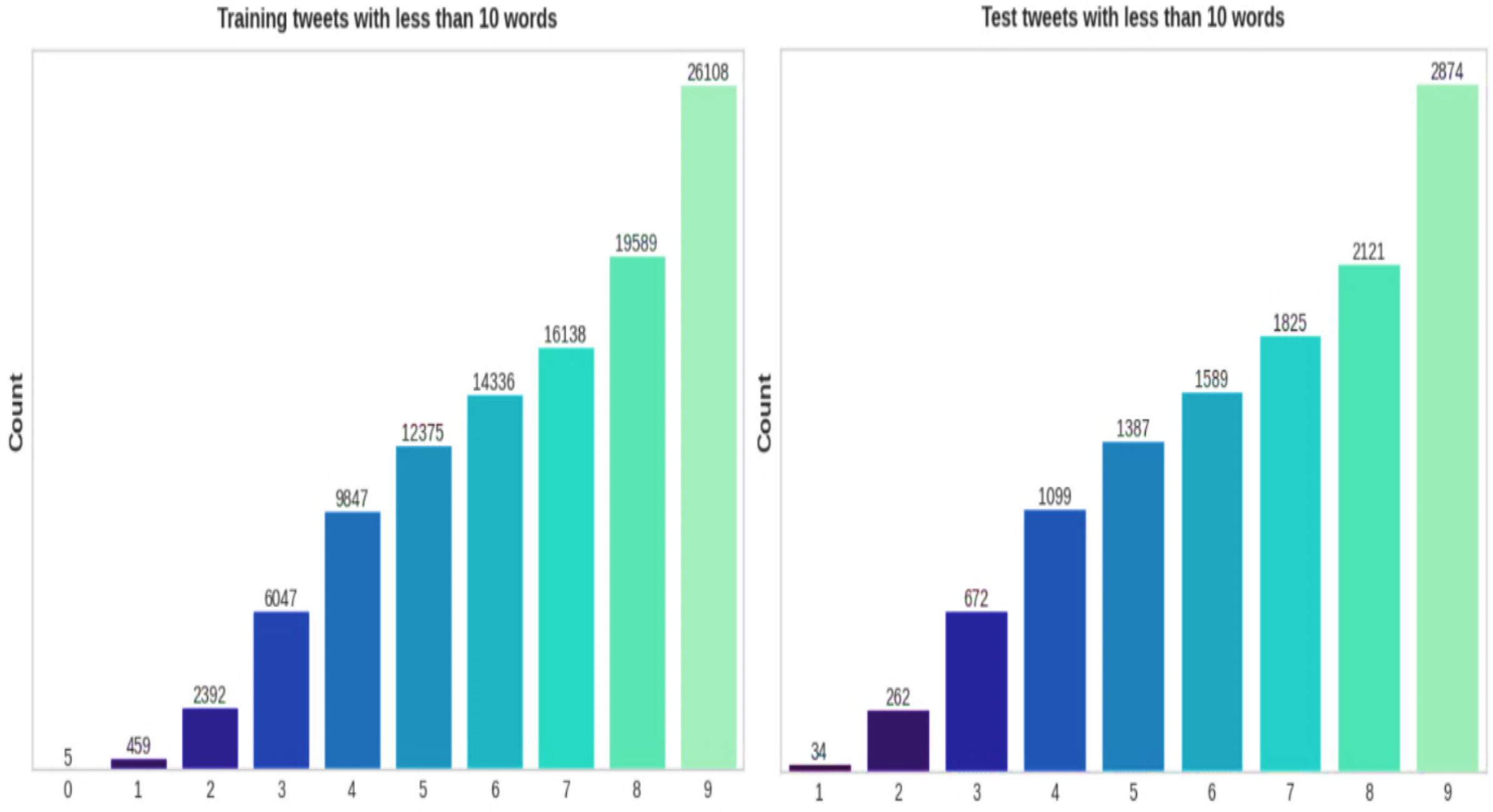

**Figure 5**: Analyzing the Effect of Text Cleaning on Tweet Length and Count

In (Fig- 5) we can see that, cleaned tweets with 0 words are present: this is because of the prior cleaning that was done. This means that some tweets contained only mentions, hashtags, and links, which have been removed. We dropped these empty tweets and also those with less than 5 words.

Later on (especially for DL models), we conducted further cleaning procedures, with tokenization and padding which is a crucial step. For a particular DL model, we utilized the particular tokenizer from its particular library, and we converted the text into tokens, which serve as the fundamental units of text for the model's processing. And, padding ensures that all input sequences to the model have the same length. DL models (e.g., LSTMs) require inputs to be of uniform length. If sequences are shorter than the maximum length, they are padded with zeros; if they are longer, they are truncated. For the BERT model, we instantiated the BERT tokenizer from the Hugging Face library using the 'bert-base-uncased' model. The model converts text to lowercase (accurately) and segments it into tokens. Iterating through cleaned data helps determine token length distribution. The maximum token length is computed to capture most text information without truncation and recorded for reference. Moreover, we identified text entries exceeding a predefined token length threshold, such as 60 tokens. This involved iterating through the data, encoding each entry, and identifying and printing the index and text of entries exceeding the particular threshold. The analysis revealed that many long tokenized sentences (those with more than 60 tokens) were not in English, indicating the need to discard these non-English sentences from the dataset, and we dropped those sentences.

```
INDEX: 54495, TEXT: sfb9gxtfzuegc4i isaacsheu b0ftwqysndb72cm xiaojiuxiaojiu olympics ittfworld thats best kind attitude
INDEX: 55416, TEXT: abhinav321g parthiv59004 amethi wala robaroo ashishnarayans8 punterlife1 ramcharan mass patil incspeaks
INDEX: 55949, TEXT: srinutheprince raajuprince rebelliontweet insoucianttgirl sampathtw zameersayed sreekanth0827 pavanspeaks
INDEX: 56387, TEXT: saurabhkv44 mohangowda96 boriamajumdar satwiksairaj shettychirag04 ha olympics aise bohot gadbad hota hai everytime
INDEX: 57920, TEXT: soccerzela 1 khune 2 hlanti 3 ngezana 4 ngcobo 5 mathoho 6 blom 8 thibedi 7 billiat 9 nurkovic 10 manga
INDEX: 58645, TEXT: amazingabhi18 navdeepsaini96 mdsirajofficial sundarwashi5 devdpd07 yuzi chahal sachinbabyy bro bad olympics
INDEX: 60384, TEXT: shivammew2005 rawatrahul9 indiatoday agr hum isi trh pure 4 saal unko support krte shyd kuch fark pad skta th
INDEX: 60559, TEXT: panhandleexit colemarisa49 booksone4 isafeyet tj2020landslide dorismele debb65762723 michel78118339
INDEX: 61526, TEXT: china advances womens quarters edge australia 7674 798cf242abd45884b3a8bcb4e5e2dcf1 foxnewssportsolympics
INDEX: 61732, TEXT: orldpeacedrea2 bimmerl0ver catsaysnyaa jerkin29813636 pplfuture boycotthegemony mfggearofficial solomonyue
INDEX: 63142, TEXT: vjones84280022 lillianna277 leslie 2348 kathysmith2115 mesinger67 marilynnv423 decontib60de missyblissey24
```

These sentences are not in english. They should be dropped.

Finally, we got the deep-cleaned dataset. The top 10 rows of the dataset, shown in Table 3.

| | text | sentiment |
|---|---|---|
| **0** | rejected sports 2020 olympics include bowling chess tug war | 0 |
| **1** | team usa men basketball team playing horrible olympics going 2004 make far | 0 |
| **2** | my_hive_away im watching kayaking olympics broadcast cool sport enjoy watching | 1 |
| **3** | girls age still trying figuring open account onlyfans 18 years old anastasija zo | 1 |
| **4** | omensfromeden patty mills olympics | 0 |
| **5** | one fav mario party mini games olympics | 0 |
| **6** | leannewhittle1 derventioexcel jacobwhittle11 olympics teamgb incredible swim proud parent day | 1 |
| **7** | becauseimatter agree feels like focus elsewhere meaning excitement olympic | 0 |
| **8** | ms dhoni shooting gun olympics vachestundi | 0 |
| **9** | whenever watch victory ceremonies years olympics know official music composed | 0 |

**Table 3:** Top Ten Rows of the Dataset

## 3.3 Data Visualization:

In DL, data visualization is crucial for deciphering complex data, seeing trends, and coming to critical conclusions. By simplifying and visualizing, decision-makers may make informed choices more quickly and accurately. A few various methods for data visualization have been used.

- *Waffle Chart:*

We used a Waffle Chart (Fig-6) to provide a visually engaging representation of the sentiment distribution. The Waffle class from the Pywaffle library is used to generate a grid-based data visualization that resembles a waffle or checkerboard, whereas the bar plot utilizes Seaborn for plotting.

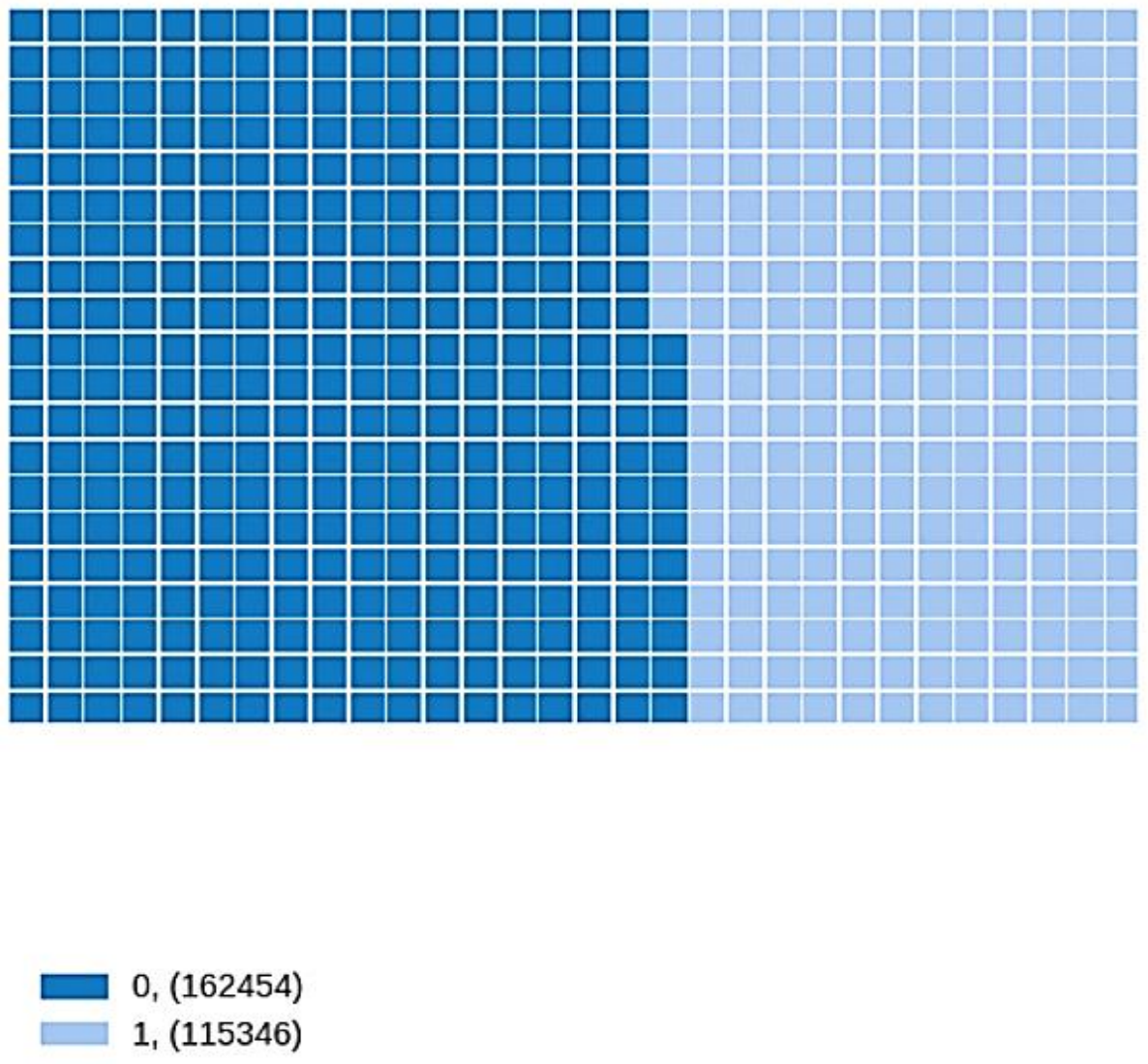

**Figure 6**: Waffle Chart

- *Count Plot:*

Count Plot is a type of bar plot used to visualize the frequency of unique values within a categorical variable, providing a simple and effective way to understand the distribution of data. (Fig- 7) count plot represents the distribution of the data, allowing us to see both positive (1) and negative (0) sentiments.

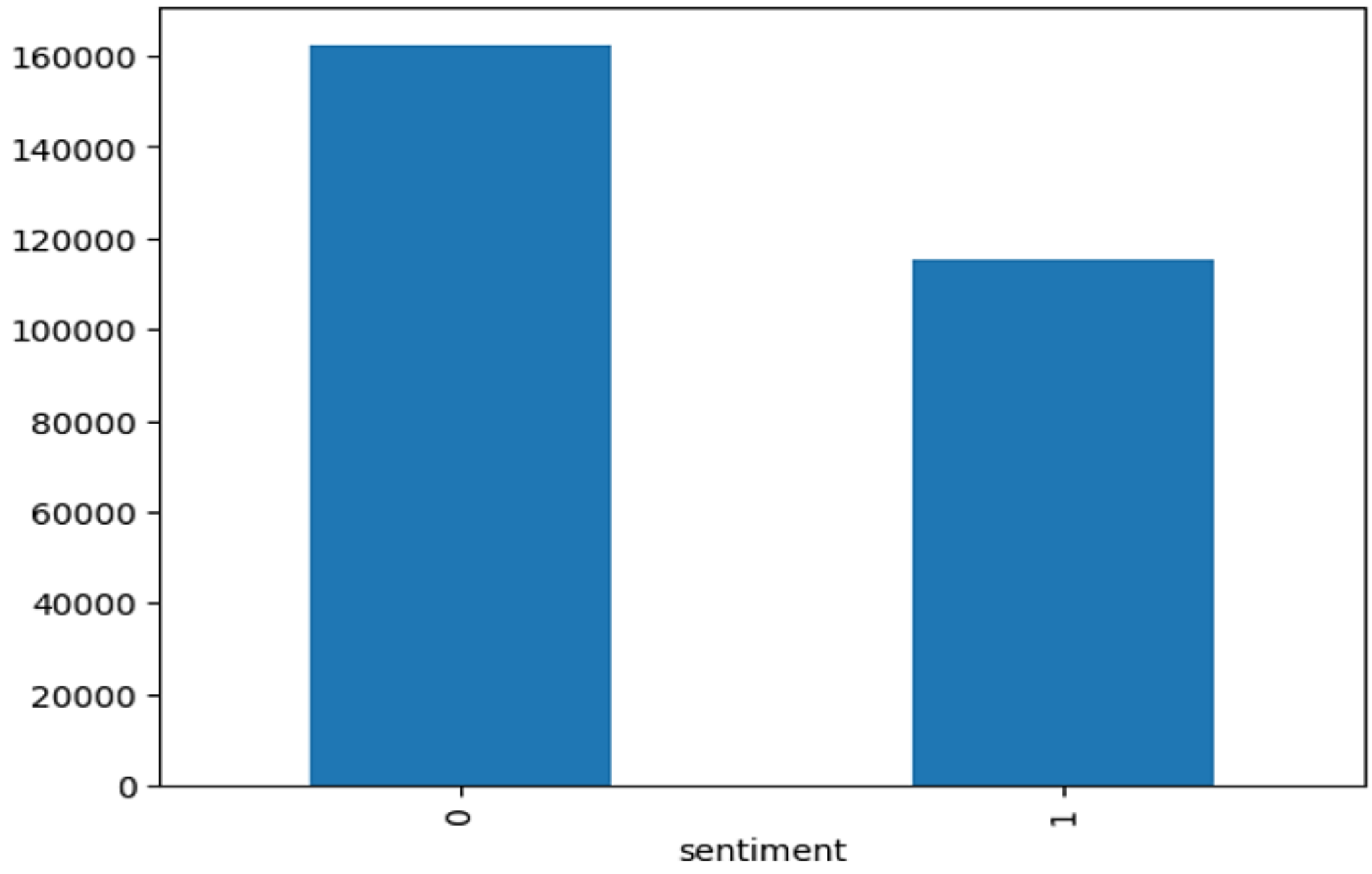

**Figure 7**: Count Plot

- *Distribution of Text Lengths:*

To understand text characteristics, we analyzed the distribution of text lengths (Fig. 8) in our cleaned text data. The text analysis comprised utilizing the Seaborn library for displaying text length distribution, constructing a visually appealing histogram with 50 bins, and flattening the distribution with a Kernel Density Estimate (KDE) curve.

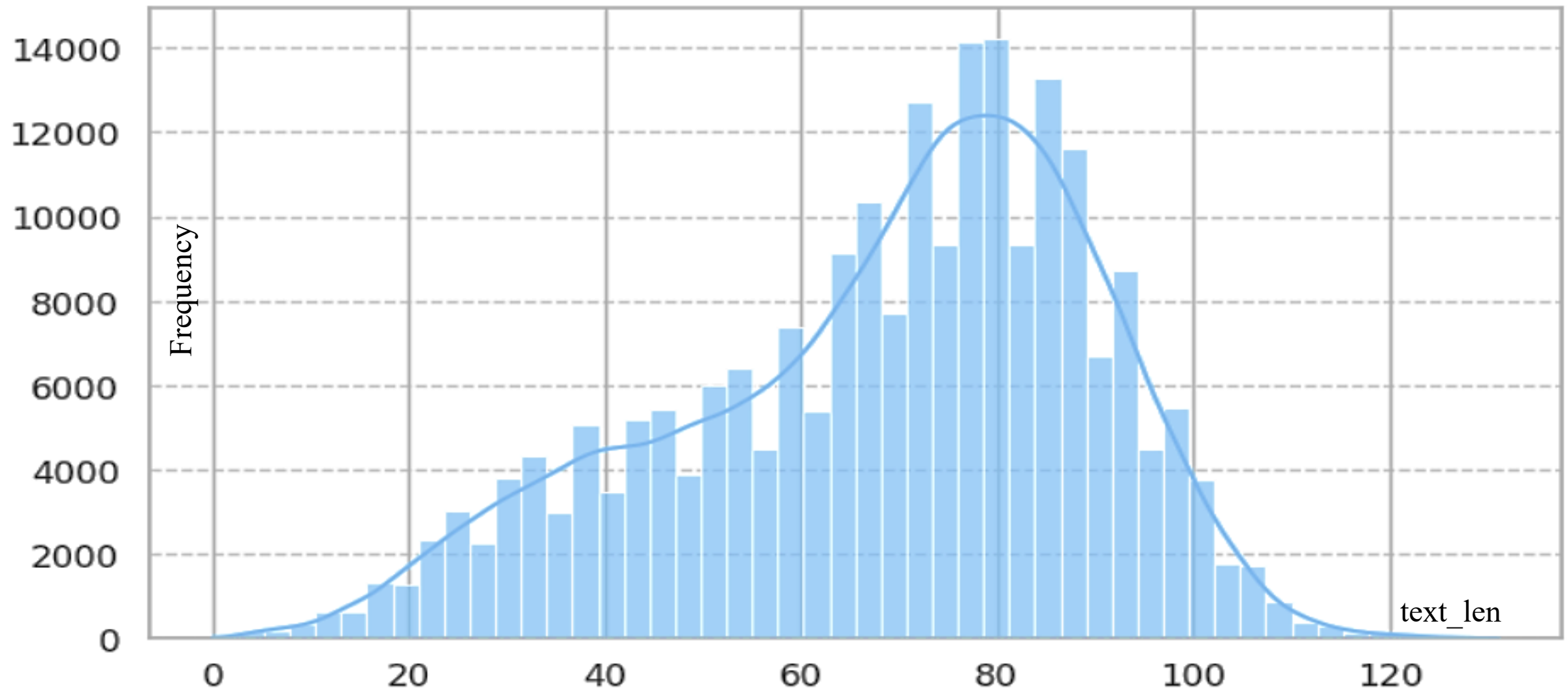

**Figure 8**: Distribution of Text Length

This analysis not only sheds light on textual diversity, but it also helps to discover any abnormalities or trends that may affect subsequent analysis or modeling efforts.

- *Frequency Table:*

We also plotted a frequency table (Fig-9), generated by a natural language toolkit function called as frequency distribution, implementation is done using simply plotting standard plotting functions. By this we can see useful words which people are talking about more on Twitter.
So, we can see the word itself "olympics" is popular or can say most frequent in the Twitter data.

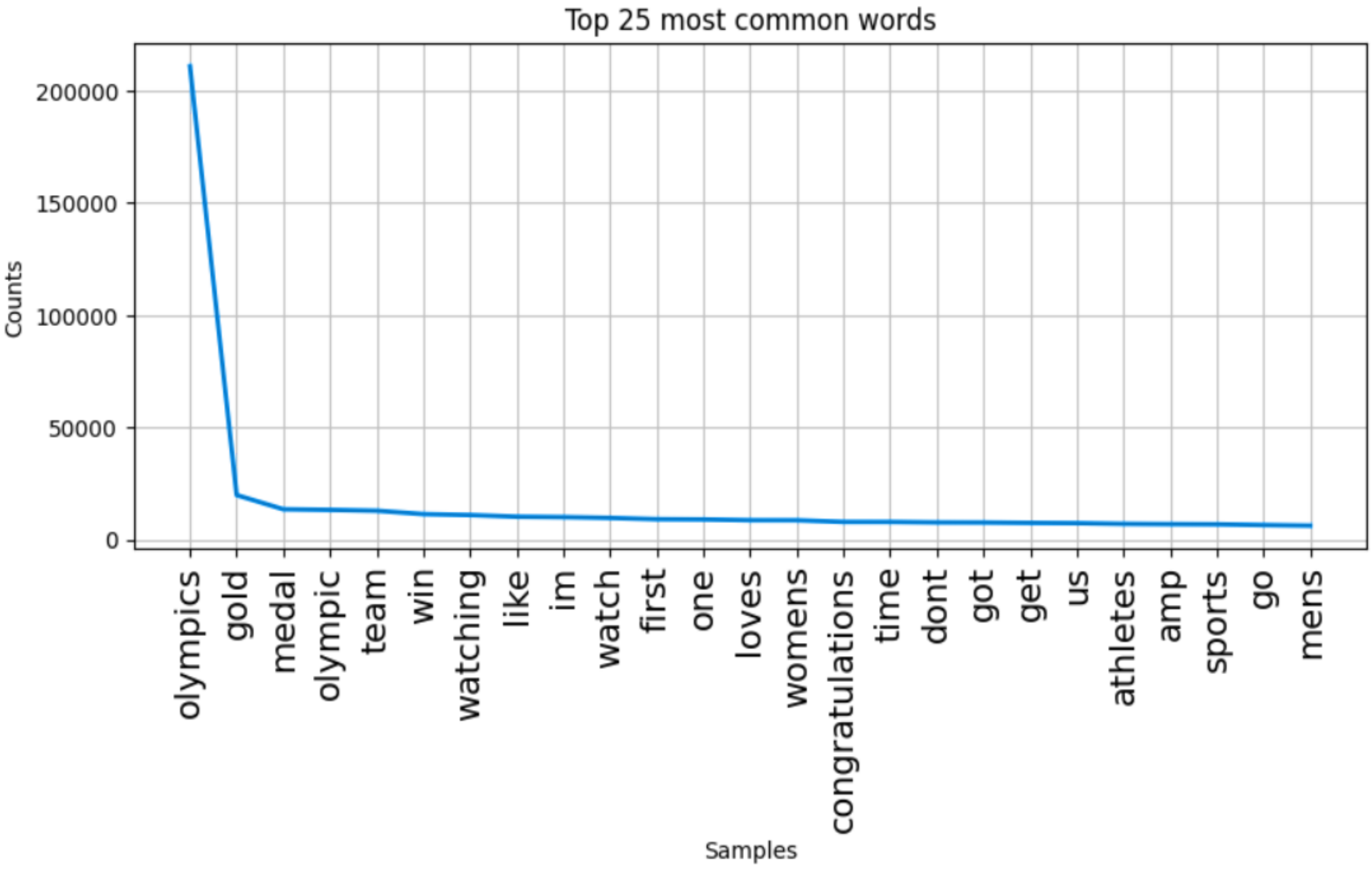

**Figure 9:** Frequency Table

- *Pie Chart:*

We also visualize our data using a pie chart (Fig- 10), we set the figure size, colours for positive and negative sentiments, defined line width and edge colour, specified tags in our pie chart and specified distance how many edges should be exploded. Then plotted our pie chart.

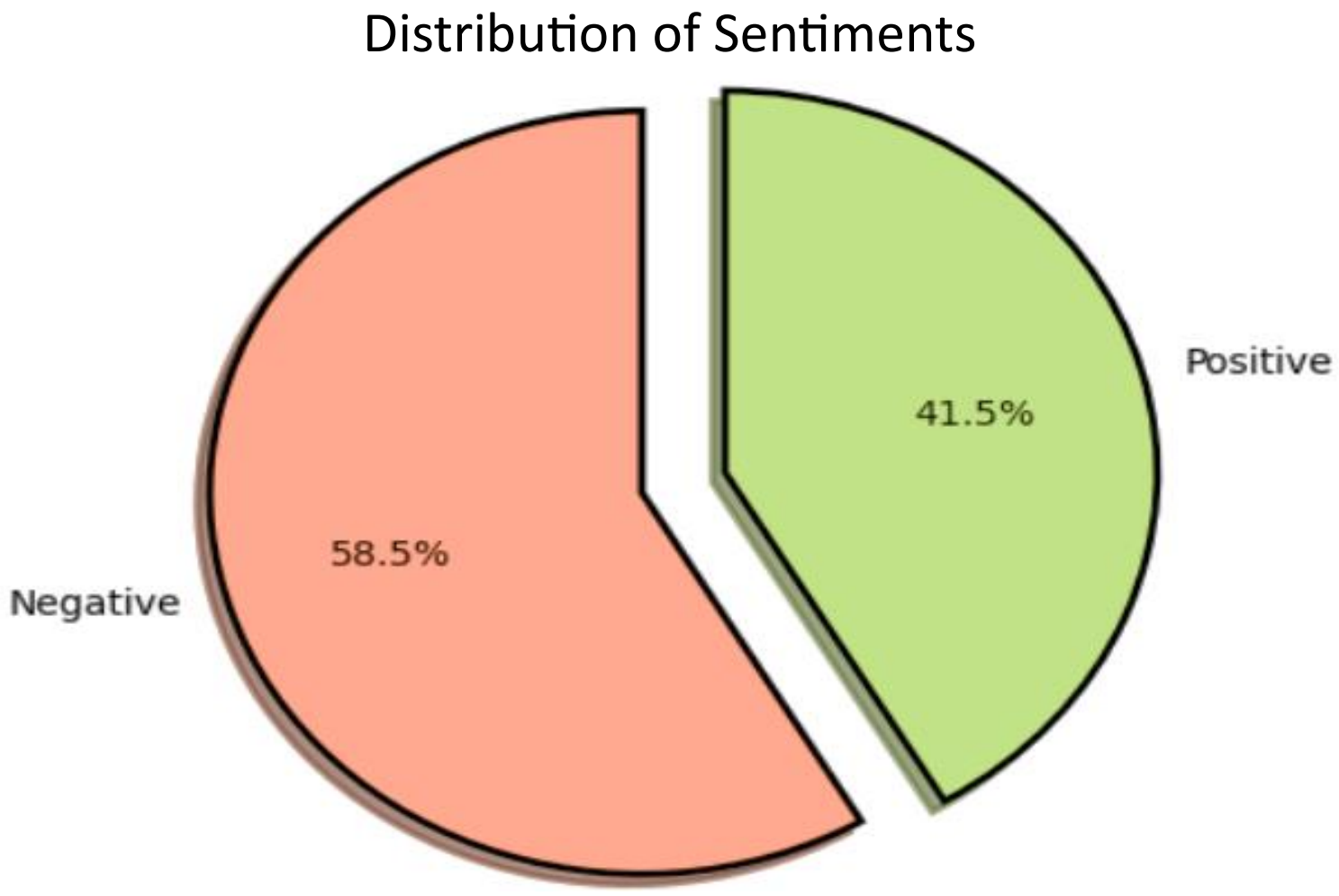

**Figure 10:** Pie Chart

With this pie chart, we can easily visualize the positive and negative sentiment distribution.

- *Word Clouds:*

We've used Word Clouds also, it's a visualization tool for text and is mainly used to visualize the words with high frequency or importance in a text. Word clouds are visually appealing and can be used for trend spotting, text summarizing, and comparative analysis. The colour and size of each word in the word cloud will vary. The words that appear the most frequently are the ones with the largest font sizes.

The word cloud 1 (Fig- 11) for only positive tweets:

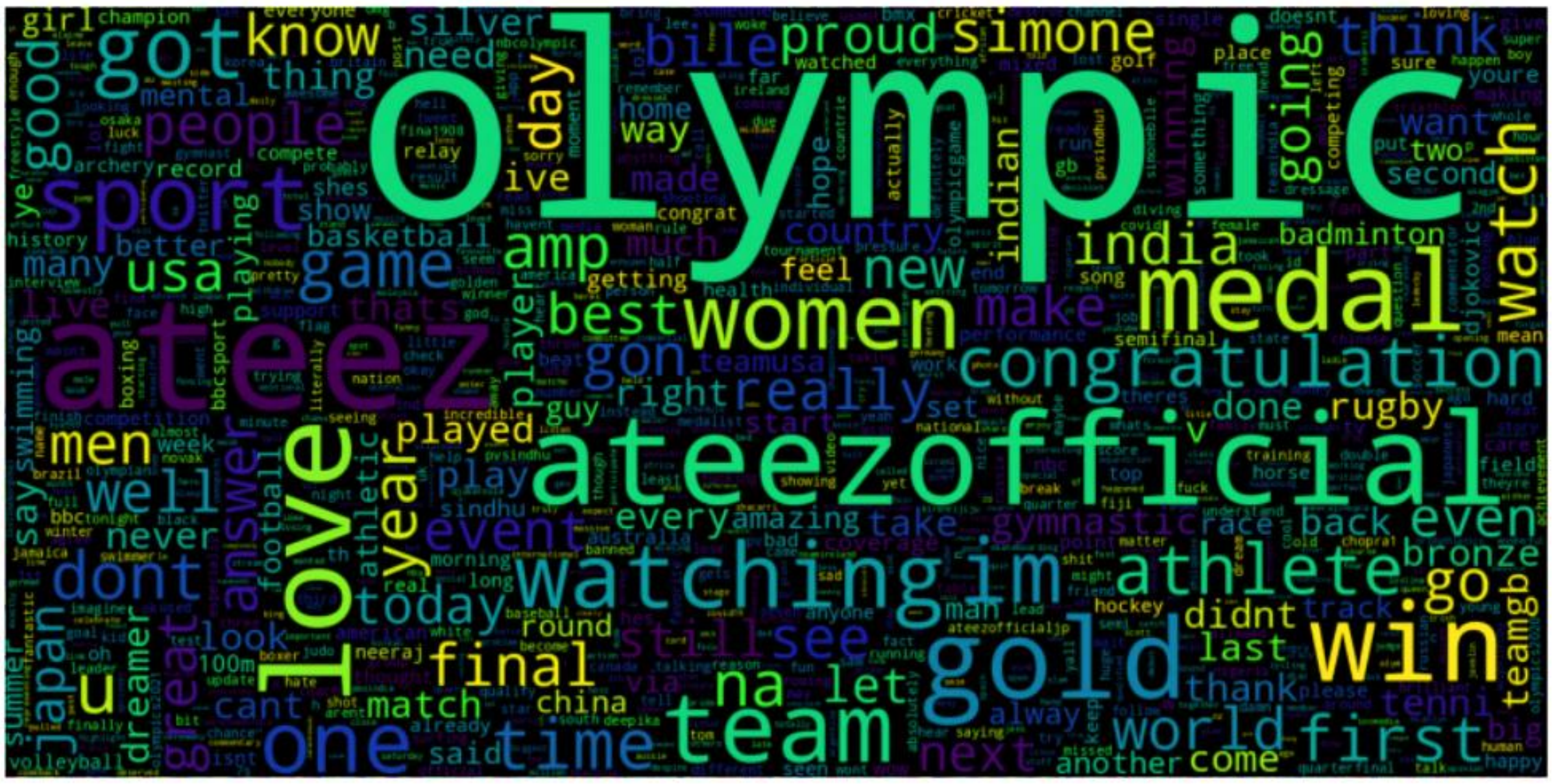

**Figure 11:** Word Cloud 1

We have categorized tweets into positive and negative categories to understand key terms and themes. Word clouds were generated for each group, allowing visual comparison of prominent words within each sentiment category.

The word cloud 2 (Fig- 12) for only negative tweets:

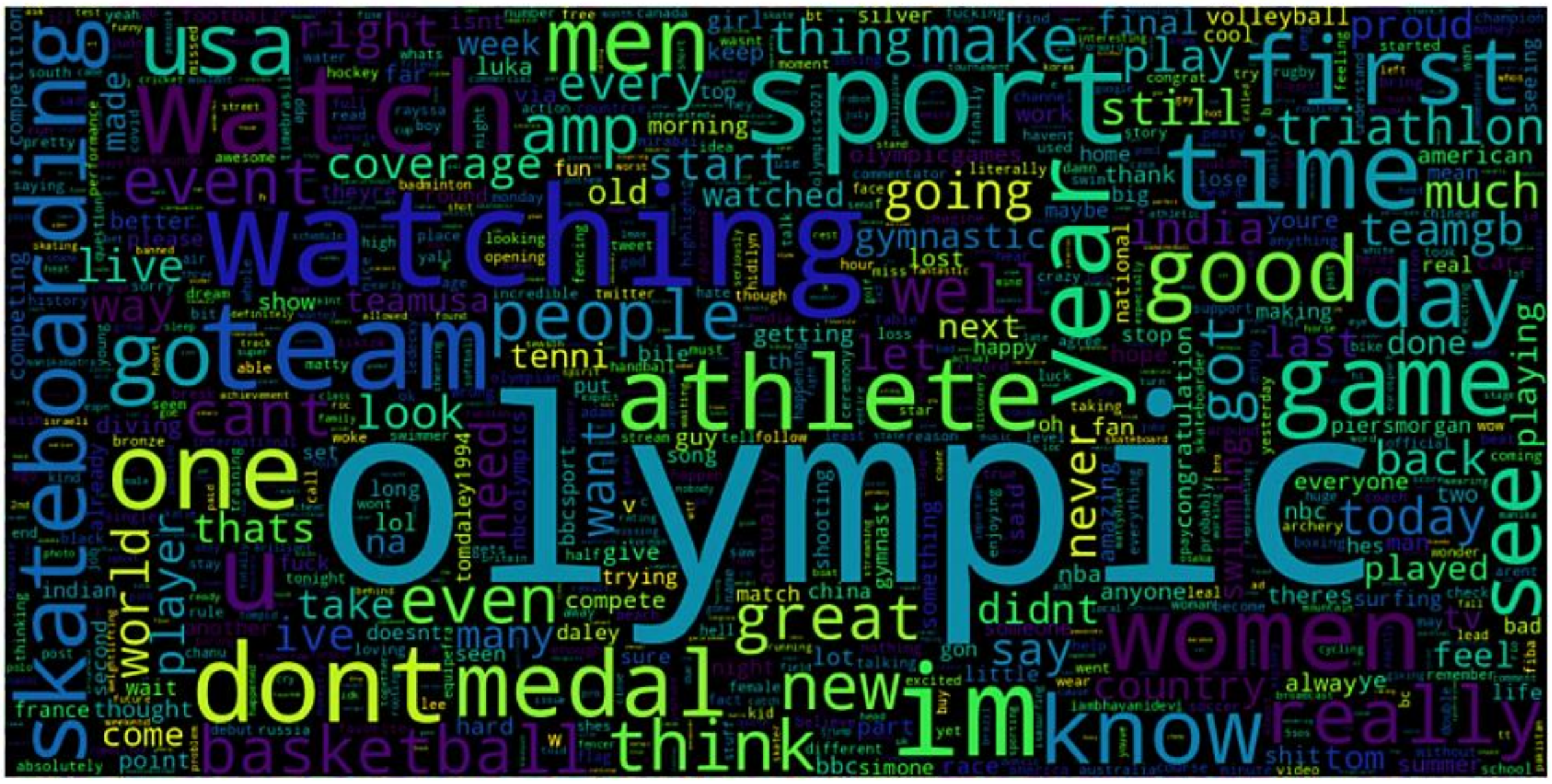

**Figure 12:** Word Cloud 2

This shows that the most frequent words in the positive tweets word cloud are "Olympic," "Ateez," "Love," "Gold," and "Medal," while in the negative tweets word cloud are "Olympic," "Watching," "Athlete," "Game," "Team."

- *Distribution of Character Lengths and Word Counts (pos & neg separated):*

Next, to do a more thorough analysis of the text data, we generated two sets of subplots. Deeper insights into the structure, as well as characteristics of the dataset based on sentiment, are made possible by these visualizations (Fig- 13), which provide an extensive understanding of the distribution of word counts and character lengths in the text data.

The distribution of character lengths for text data categorized by sentiment can be observed in the first set of histograms we generated. These histograms show the frequency of character lengths within each sentiment category and distinguish texts with positive and negative sentiments.

In the second set of subplots, we illustrated based on sentiment, the word count distribution in the text data. Similar to the first set, these histograms distinguish between texts with positive and negative sentiments, revealing the distribution of word counts within each sentiment category.

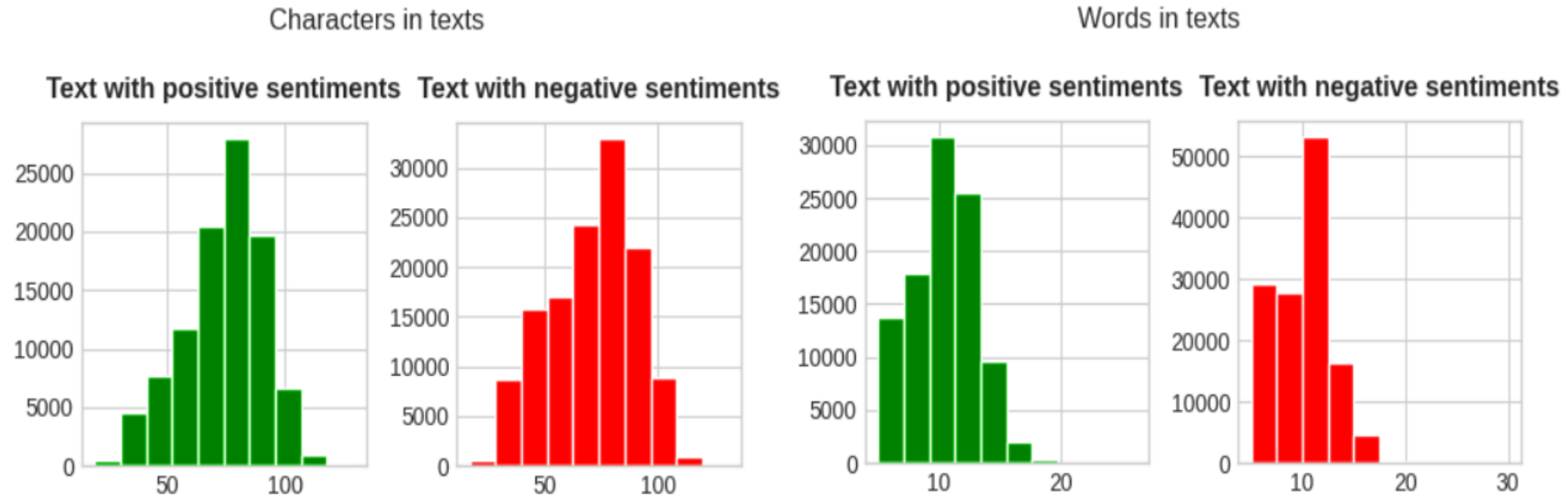

**Figure 13:** The Distribution of Character Lengths and Word Counts in the Text data

- *Relationship b/w Sentiment & Avg. Word Length:*

We have furthermore analyzed textual data to explore the relationship between average word length and sentiment (Fig. 14). We tokenized the text into words and determined the length of each word in order to get the average word length. Next, for every text sample, we calculated the average word length. Using a density plot (shown in green), we were able to see the distribution of average word lengths for texts with positive emotion. Similarly, we plotted the distribution of sentences that were negative in red. The produced graphs shed light on how texts with positive and negative sentiments differ in terms of average word length.

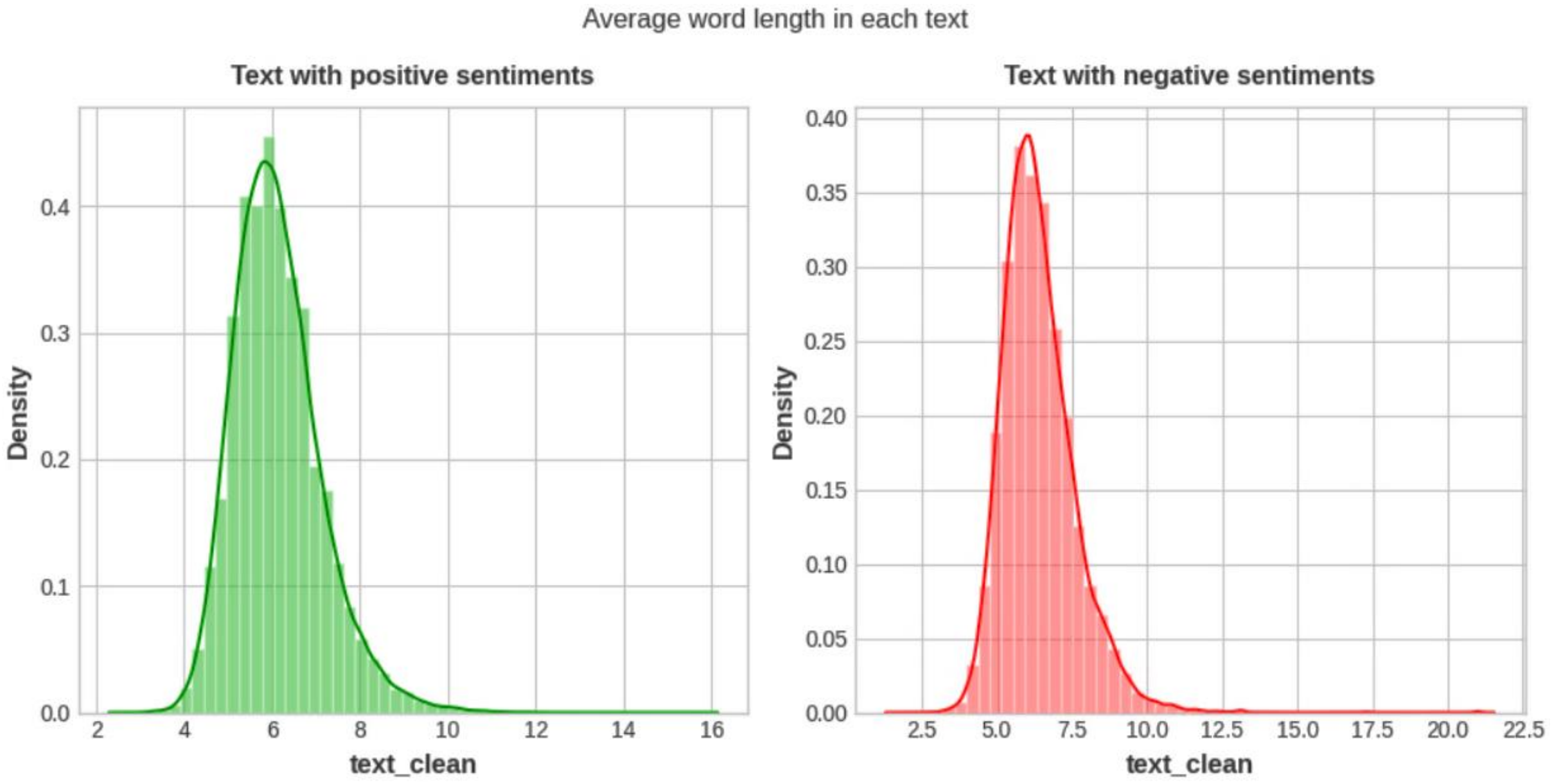

**Figure 14:** The Relationship between Sentiment and Average Word Length

### 3.4 Feature Extraction and Visualization:

In this study, we don't have to use feature selection because the relevant feature is already included in the dataset. Feature selection is often used to remove redundant or unnecessary variables. The process of feature extraction involves converting textual input into real-valued vectors. Text data is not understood by the computer so it's necessary to convert the text data to a numerical value, then it's compatible to fit in the ML/DL model.

- *Count Vectorizer and TF-IDF:*

We employed two commonly used models for this purpose: Count Vectorizer (CV) and TF-IDF.
Firstly, the CV model transforms raw text data into a Document-Term Matrix (DTM) [44], portraying each column as a distinct word in the vocabulary and every row as a document. This results in a sparse matrix, with most elements being zero. In a dataset of 2,77,800 tweets, 1,64,363 unique features (words) were observed, but the percentage of non-zero elements was low. The CV maps words from the vocabulary to their occurrences in each document, aiding in identifying word patterns associated with different documents or classes.
Let $D$ be the set of documents, $W$ be the vocabulary (set of distinct words), and $nij$ be the count of word $j$ in document $i$.
The CV transforms the raw text data into a DTM $X$ where each element $xij$ signifies the count of word $j$ in document $i$.

$$xij=nij$$

The size of the DTM $X$ is $|D|\times|W|$, where $|D|$ is the quantity of documents and $|W|$ is the size of the vocabulary [45].

Then, Term frequency-inverse document frequency (TF-IDF), is used for feature extraction also. For information retrieval, it is a statistical metric basically used in NLP to assess the significance of a word in a document relative to a set of documents.
Let, $TFij$ be the term frequency of word $j$ in document $i$, $DFj$ be the document frequency of word $j$ (i.e., the number of documents containing word $j$), and $N$ be the total number of documents.
Term Frequency ($TFij$) is the frequency of word $j$ in document $i$ normalized by the total number of words in document $i$. It is calculated as:

$$TF_{ij} = \frac{n_{ij}}{\sum_k n_{ik}}$$

Inverse Document Frequency ($IDFj$) is the inverse fraction (logarithmically scaled) of the documents, containing the word $j$. It is calculated as:

$$IDF_j = \log\left(\frac{N}{DF_j}\right)$$

The TF-IDF score ($TFIDFij$) for word $j$ in document $i$ is the product of $TFij$ and $IDFj$:

$$TFIDF_{ij} = TF_{ij} \times IDF_j$$

The TF-IDF transformation yields a matrix $X'$ where each element $xij'$ represents the TF-IDF score [46].

Using CV and TF-IDF both, this combination enhances model performance by capturing raw term frequencies, penalizing common words and boosting rare ones, normalizing frequencies, improving discriminative power, reducing common word impact, and maintaining sparsity for better feature representation in downstream models, making it a more nuanced approach.

- *N-gram models:*

In our exploration of feature visualization using N-gram models for SA, we investigated the effectiveness of combining unigrams and bigrams as features. We actually applied CV to extract features while considering both single words (unigrams) (Fig- 15) and pairs of consecutive words (bigrams) (Fig- 16). Then, a Naïve Bayes (NB) classifier was employed as an external baseline classifier to model the relationship between the sentiment labels and the retrieved features. The entire pipeline, including feature extraction and classification, was contained in a 'Pipeline' object to facilitate rapid execution. To better comprehend the model's most influential features, we visualized the coefficients associated with the top 25 features.

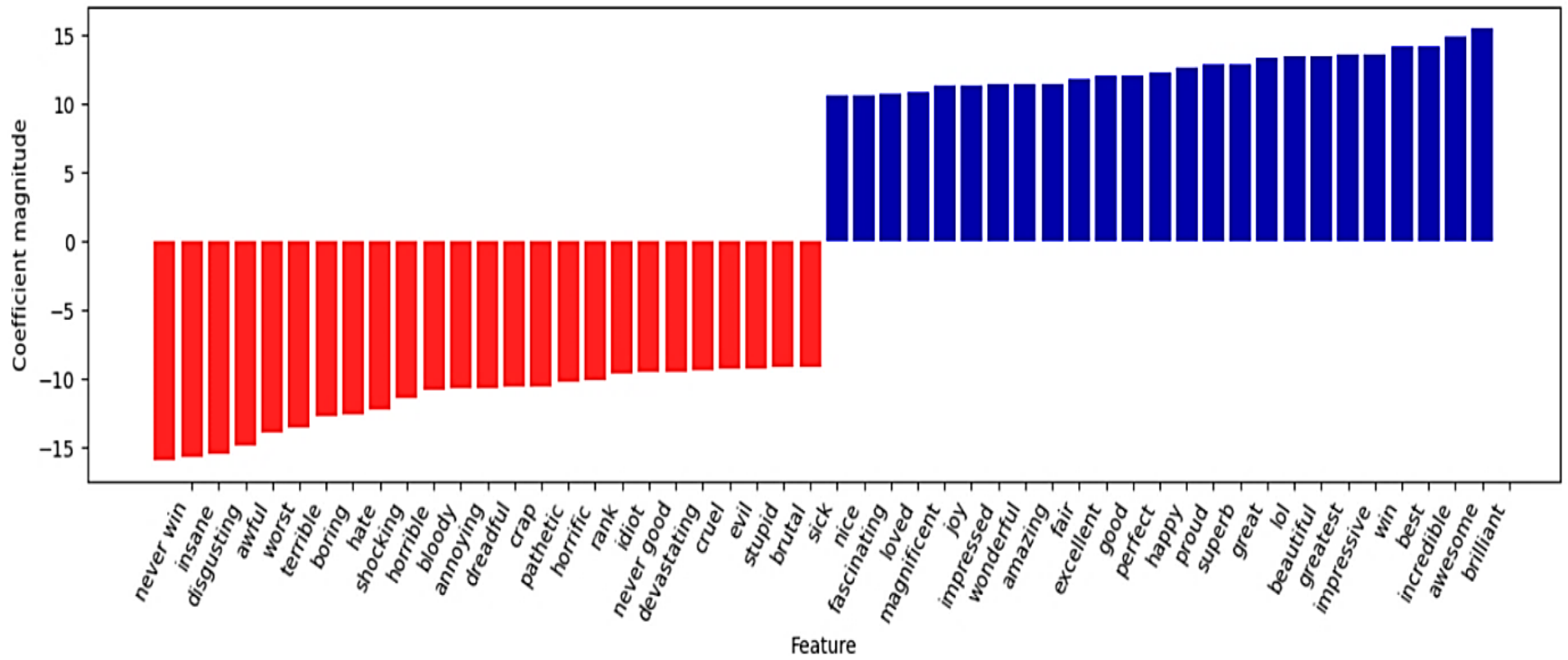

**Figure 15:** Unigram Features

These are single words, as individual features. Each word in the text corpus is considered independently.

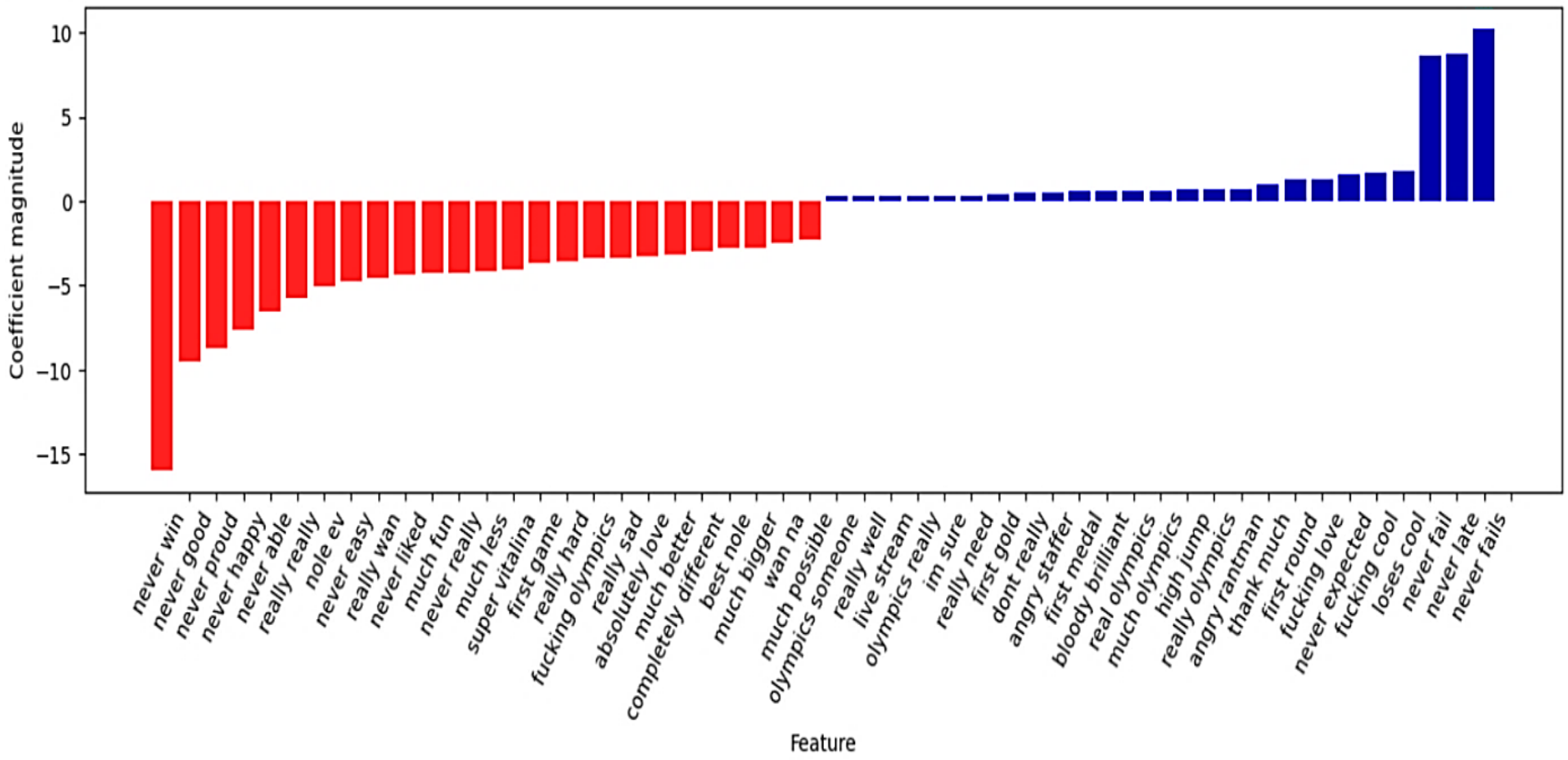

**Figure 16:** Bigram Features

These are pairs of consecutive words treated as features. Bigrams capture the co-occurrence of words.

The most significant unigrams and bigrams influencing sentiment categorization are shown in this feature visualization. The exploration of N-gram models, coupled with the visualization of feature importance, offers valuable understanding of the relationship between textual features and sentiment labels. This contributes significantly to the field of SA methodologies, enhancing our comprehension of how specific word combinations impact sentiment prediction.

• *Word embeddings*:
Word embeddings were also implemented especially for our advance DL models. One of the most often used ways of representing document vocabulary is word embedding, which can capture a word's relationship to other words, the context inside a document, and syntactic and semantic similarity [47], etc. Word embeddings, a form of dense vector representations for words, have gained prominence for their ability to capture semantic relationships, offering a richer representation compared to sparse methods.

• *Word2Vec:*

Using NNs or matrix factorization, it is possible to learn the word embeddings. Word2Vec, developed by Google is a widely used method for learning word embeddings using shallow NNs. It basically learns word embeddings from the text using a computationally efficient NN prediction model. Word2Vec can create word embeddings using two methods (Fig- 17) (both involving NNs): the Common Bag of Words (CBoW) and the Skip-Gram (SG) [48].
The SG model can predict the context words given the target word, whereas the CBOW model from context words (for example, "the man is _ chess," and the target word is denoted by "_") can predict the target word ("playing"). The CBOW model statistically smooths away a significant amount of distributional information by treating the complete context as a single observation. It works very well with small datasets [49]. But, the SG model works better with bigger datasets because it treats every context-target combination as a novel observation [50].

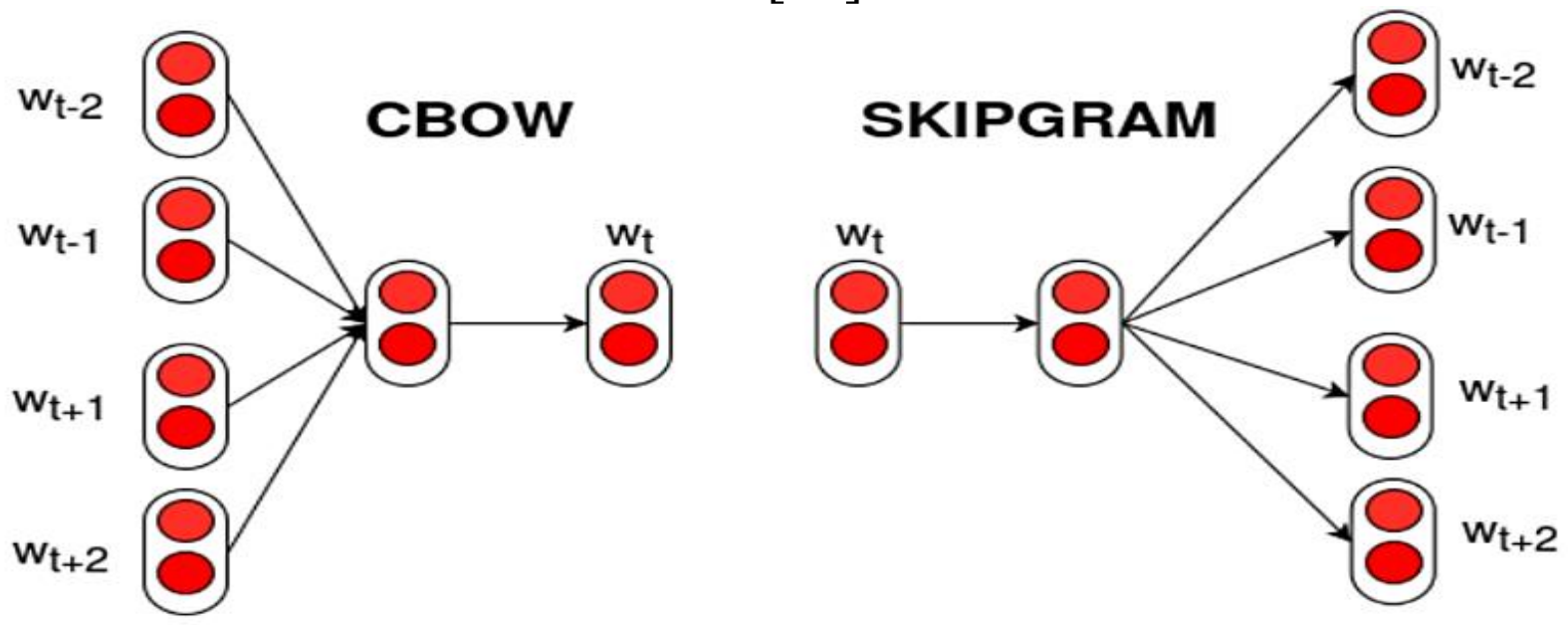

$$\frac{1}{T}\sum_{t=1}^{T}\sum_{-c\leq j\leq c, j\neq 0} logp(w_t|w_{t+j}) \qquad \frac{1}{T}\sum_{t=1}^{T}\sum_{-c\leq j\leq c, j\neq 0} logp(w_{t+j}|w_t)$$

**Figure 17:** CBOW and SG models in Word2Vec.

The Word2Vec model was implemented by tokenizing tweets, training it using Gensim, extracting vocabulary, creating document vectors, and constructing a feature matrix.

Word embeddings are effective NLP tools but have drawbacks like generating single vectors per word, lacking contextual awareness, and struggling with out-of-vocabulary words. They require a fixed vocabulary set, are computationally expensive, high dimensional, and can amplify biases. Many DL models themselves play the role of feature extractors [51].
Transformer-based models, such as BERT, facilitate transfer learning, handle polysemy, offer context-sensitive embeddings, and deal with multimodal data. The BERT opened the path for more sophisticated and context-aware models in modern NLP applications.

## 3.4 Dataset Splitting:

Data splitting is a crucial process that involves dividing a dataset into distinct subsets (e.g., training, validation, and testing). The BiLSTM model uses the cleaned dataset (100%) split into three sets: 70% for training, 15% for validation, and 15% for testing. The train_dataset (90%) is split into training and validation sets in an 80-20 ratio, 72% of the total data is used for training and 18% is used for validation for both the CNN and BERT models. The test_dataset (10%) remains unchanged. To avoid prolonged training times, only for the CNN model, a smaller subset of data is used initially, with the first 50,000 entries taken for validation and the remaining for training.

### 3.5 Model Building:

- *Baseline Model Building:*

In the field of SA utilizing DL, baseline models are an essential component in assessing the efficacy and performance of more complex NN architectures. Baseline models are essential for creating a standard by which the improvements of complex models are evaluated. They are distinguished by their simplicity and ease of use.
We used the NB model as our baseline classifier in our research project. This model, which is well known for being simple to learn and easy to use, served as a base for SA before DL approaches were explored. The feature representation used to train the NB classifier is derived from traditional feature extraction techniques like TF-IDF and CV.

  - *NB:*

Based on the Bayes' theorem, the NB model is a probabilistic classifier [52], it determines the likelihood of a document belonging to a specific class by combining the probabilities of individual features. It is computationally efficient, especially with high-dimensional feature spaces and sparse data [53]. However, there are certain limitations, such as assuming feature independence and capturing complicated interactions. Overall, the NB model offers useful insights and predictions for textual data analysis. The efficacy of DL models in SA tasks is evaluated by comparing their complexity to baseline models, to establish the best strategy for real-world applications.

- *Advance Model Building:*
  - *CNN:*

CNNs, while generally employed for image processing, have also proven effective in text classification tasks such as SA. In this context, CNNs use convolutional filters to capture local features from text sequences. By sliding these filters across word embeddings, CNNs may discover sentimental patterns such as phrases or n-grams. Afterward, pooling layers aid in dimensionality reduction and feature highlight. This method enables CNNs to learn and emphasize significant sentiment-carrying phrases, making them useful for text categorization [54].
CNN employs a multilayer perceptron variant designed to require minimal pre-processing. CNN is a type of deep, feed-forward ANN (whereby the connection of the nodes doesn't create a cycle). These are inspired by the animal visual cortex [55].
The following describes the configuration and methodology used.

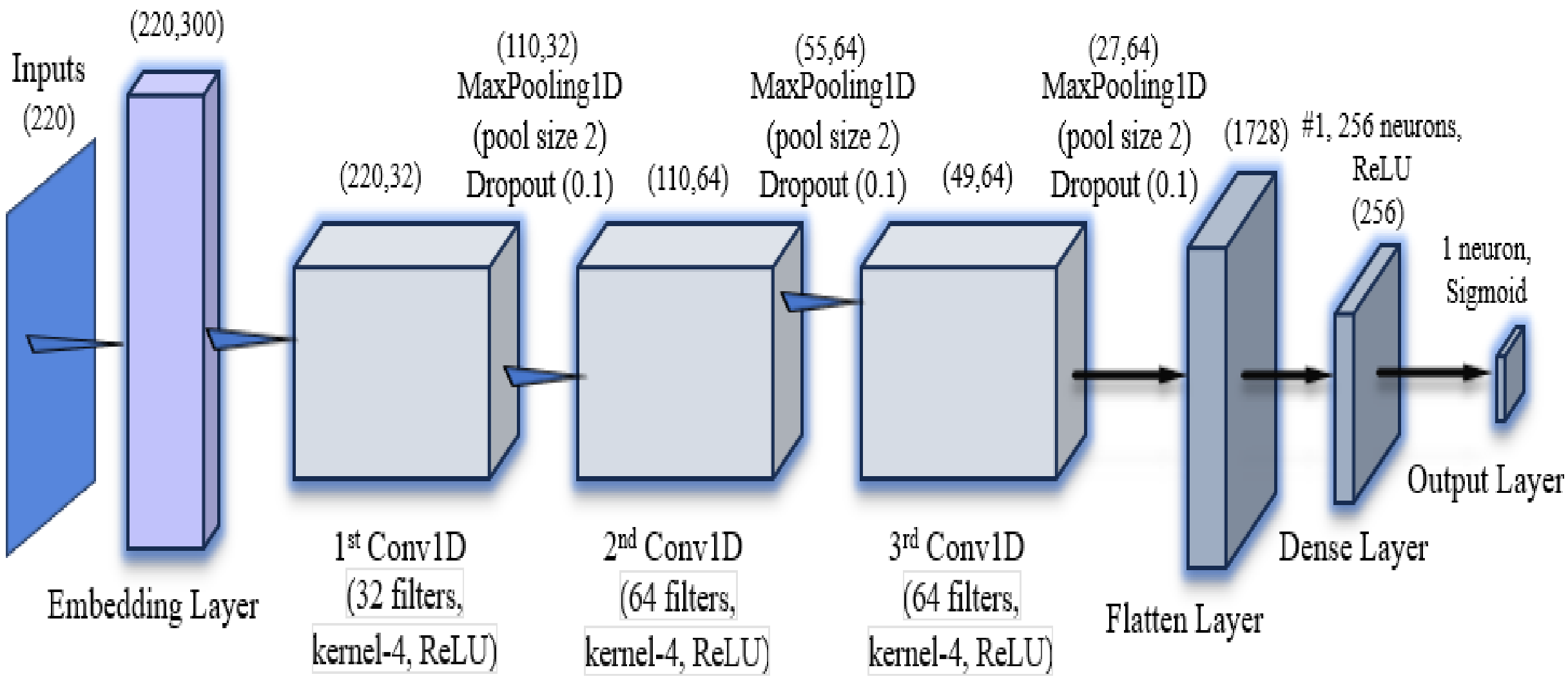

**Figure 18:** Proposed CNN Model Framework

A 300-dimensional vector space (defined by the embedding size (EMBED_SIZE)) containing unique words is represented by the word index length, which determines the vocabulary size (VOCAB_SIZE). The CNN model is optimized by the Adam optimizer [56] (with a learning rate of 0.01), which combines the advantages of Stochastic Gradient Descent (SGD) with RMSProp [57] for enhanced training efficiency. To minimize overfitting and avoid overconfidence in predictions, the Loss Function is set up using binary cross-entropy and label smoothing (with a factor of 0.1). During training, an Early Stopping callback monitors the accuracy of the model and stops the training if there is no improvement is seen after two epochs. This helps to prevent overfitting and reduce unnecessary computation.

The CNN model is constructed sequentially with the following layers:

- → Embedding Layer: Converts words into a 300-dimensional vector space.
- → Convolutional (conv) Layers: Three 1D conv layers with filter sizes of 32, 64, and 64. Using ReLU activation and a kernel size of 4, each conv layer captures local features from the text.
- → Max Pooling Layers: A max pooling layer (with a pool size 2) comes after each convo layer to reduce the computational load and spatial dimensions.
- → Dropout Layers: To mitigate overfitting, three dropout layers with a 10% rate are added after each pooling layer. During training, 10% of the input units are randomly set to zero.
- → Flatten Layer: Converts the convo layers' 2D matrix output to a 1D vector.
- → Dense Layers: A fully linked layer with 256 neurons with ReLU activation captures complex representations. This is followed by a single neuron output layer with sigmoid activation, which performs the final binary classification (positive or negative sentiment).

- *BiLSTM:*

LSTMs, a specialized form of RNNs, Improve SA by integrating memory cells and gating mechanisms, which allow them to preserve and update information over lengthy sequences, effectively handling complex sentence structures and contexts [58].
BiLSTM, a sequence processing model made up of two LSTMs (that is the core functionality): one takes the input forward, while the other takes it backward direction. BiLSTMs efficiently increase the network's information pool, enhancing the context that the algorithm has access to (e.g., recognizing which words immediately follow and precede a word in a sentence) [59].

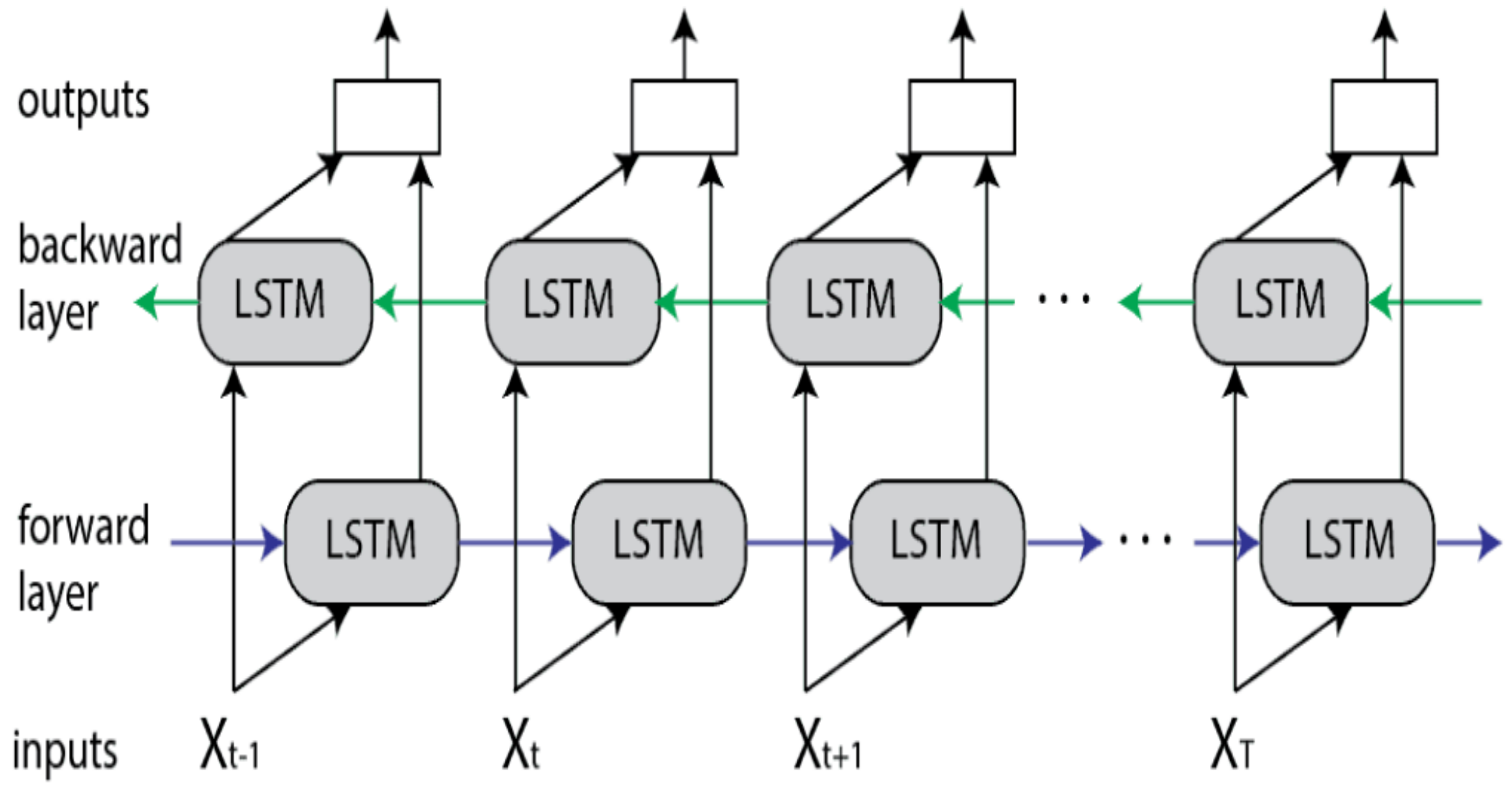

**Figure 19:** An Illustration of the BiLSTM Model

If we deep dive into the BiLSTM model it actually leverages a combination of CNN and bidirectional recurrent neural network (BiRNN) layers to classify sentiments [60].

The following describes the configuration and methodology used.

The optimizer configuration and several hyperparameters are established after importing the required libraries from Keras:

- → Vocabulary Size: The model takes into account 5000 distinct terms from the dataset.
- → Embedding Size: A 32-dimensional vector space is used to represent each word.
- → Epochs: The model undergoes 20 training epochs.
- → Learning Rate: 0.1 (initially), decay rate is derived by dividing learning rate by no. of epochs.
- → Momentum: 0.8 is used to accelerate gradients in the relevant direction and reduce oscillations.
- → SGD Optimizer: SGD optimizer is configured with the specified learning rate, momentum, decay, and without Nesterov momentum.

The BiLSTM model is constructed sequentially with the following layers:

- → Embedding Layer: use the vocabulary size (5000) to convert word indices into dense vectors of fixed size (32 dimensions).
- → Convolutional Layer: To capture local features in the text sequences, a 1D conv layer with 32 filters and a kernel size of 3 uses the same padding and ReLU activation.
- → Max Pooling Layer: Reduces the spatial dimensions (by a factor of 2), reducing computation, and preventing overfitting.
- → Bidirectional LSTM Layer: Incorporates a bidirectional LSTM with 32 units, allowing the model to learn dependencies in both (forward and backward) directions of the text sequence, thereby capturing more contextual information.
- → Dropout Layer: To avoid overfitting, a dropout layer (with a rate of 0.4) is employed, randomly setting 40% of the input units to zero while training.
- → Dense Layer: The final dense layer with 2 units and softmax activation for binary classification (positive or negative sentiment).

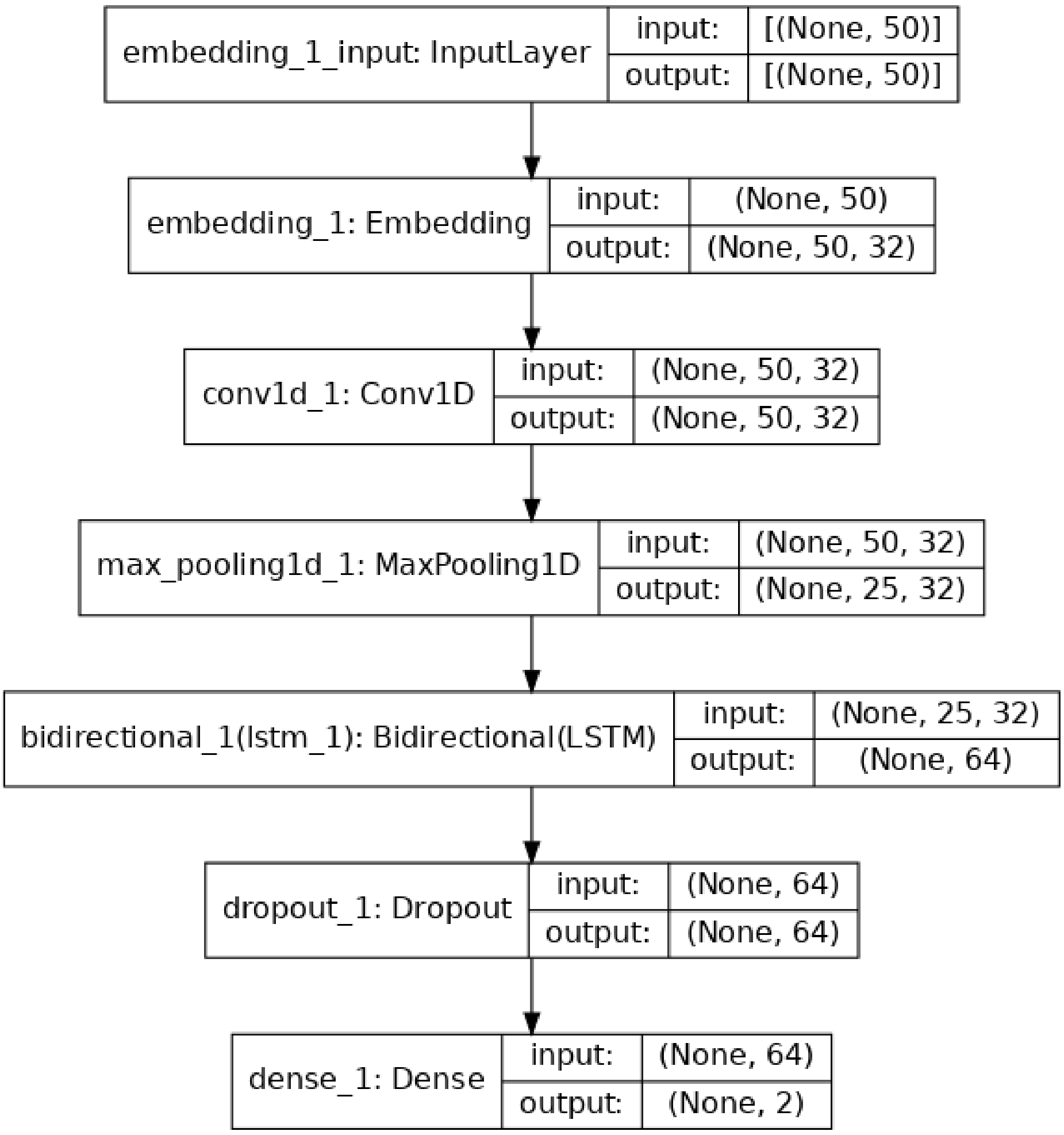

**Figure 20:** Proposed BiLSTM Model Architecture

Using the categorical cross-entropy as the loss function and a specified SDG optimizer, the BiLSTM model is compiled and then trained on over 20 epochs. The training process is verbose with a batch size of 64 and provides detailed output for each epoch.

- *BERT:*

BERT is a state-of-the-art Transformer-based model developed by Google, and the Transformer model represents a paradigm shift in NLP. In LSTM networks words are passed and generated sequentially, less effective in capturing the word's actual meaning. Even BiLSTMs learn left-to-right and right-to-left context separately before concatenating them, thus the genuine context is slightly lost. But Transformer models like BERT can process the entire text sequence simultaneously, capturing context from both directions. So, it is Deeply Bidirectional. This bidirectional understanding allows BERT to grasp the full context of each word, including nuances that are crucial for SA [61] [62].
BERT is based on an attention mechanism, the Transformer, which identifies the contextual links between words in a text. An encoder reads the input text, and a decoder generates a prediction for that task, making up a basic Transformer [63]. This system is more appealing than LSTM cells because it separates tasks, allowing for understanding English grammar, context, and language relationships.
The encoder portion is all that BERT needs because its goal is language representation model generation. If we stack only the encoders, we obtain BERT [64].

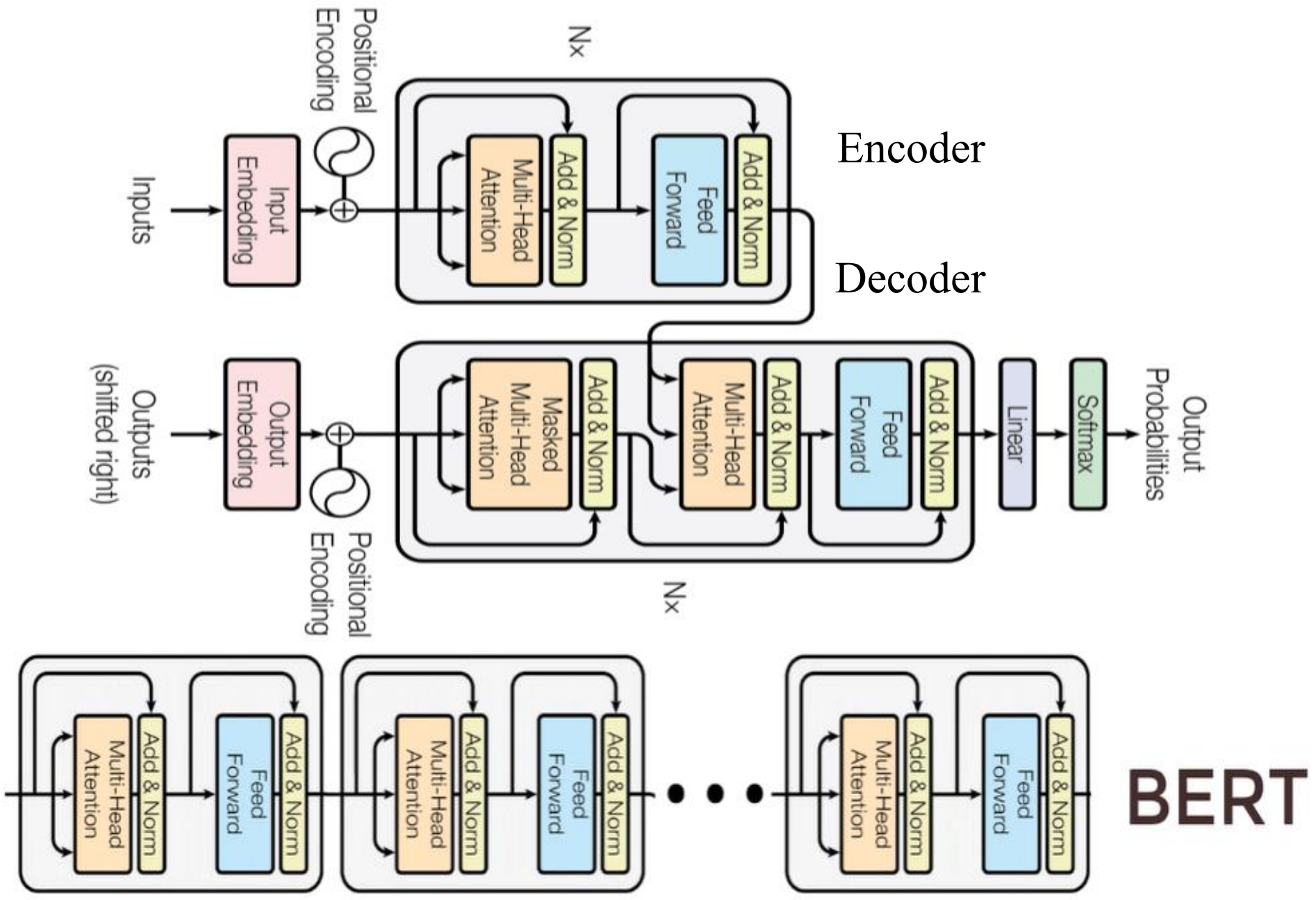

**Figure 21:** Transformer Flow and BERT

Pre-training and fine-tuning are the two main phases of BERT training. Through two unsupervised tasks, masked language modeling (MLM) and next sentence prediction (NSP), the pre-training phase aims to teach BERT the structure and context of language. In MLM, BERT predicts masked words within sentences to help it grasp the bidirectional context. BERT improves its understanding of inter-sentence relationships in NSP by determining if one sentence follows another. Inputs are sentence pairs containing masked words that are converted into embeddings using pre-trained word piece embeddings. Word vectors for MLM and a binary classification for NSP are the outputs of the model.
BERT is fine-tuned for specific NLP tasks such as question answering, replacing output layers with task-specific ones, and performing supervised training on relevant datasets. Token, segment, and position embeddings are among the input embeddings that maintain context and order. By focusing on masked words and maintaining context and order, cross-entropy loss is utilized to enhance context understanding [65].
Models, like BERT Base with 110 million parameters, achieve pretty much high accuracy.

The BERT encoder takes a series of tokens as input, which are converted into vectors before being processed by the NN. However, before processing begins, BERT demands that the input be cleaned up with some extra metadata:

- → Token embeddings: First, a [CLS] token is appended to the input word tokens at the start of the sentence, and each sentence ends with a [SEP] token.
- → Segment embeddings: Each token is added with a marker that indicates whether it is Sentence A or B. The encoder can now distinguish between sentences.
- → Positional embeddings: Every token has a positional embedding added to it to show where it is in the text [66].

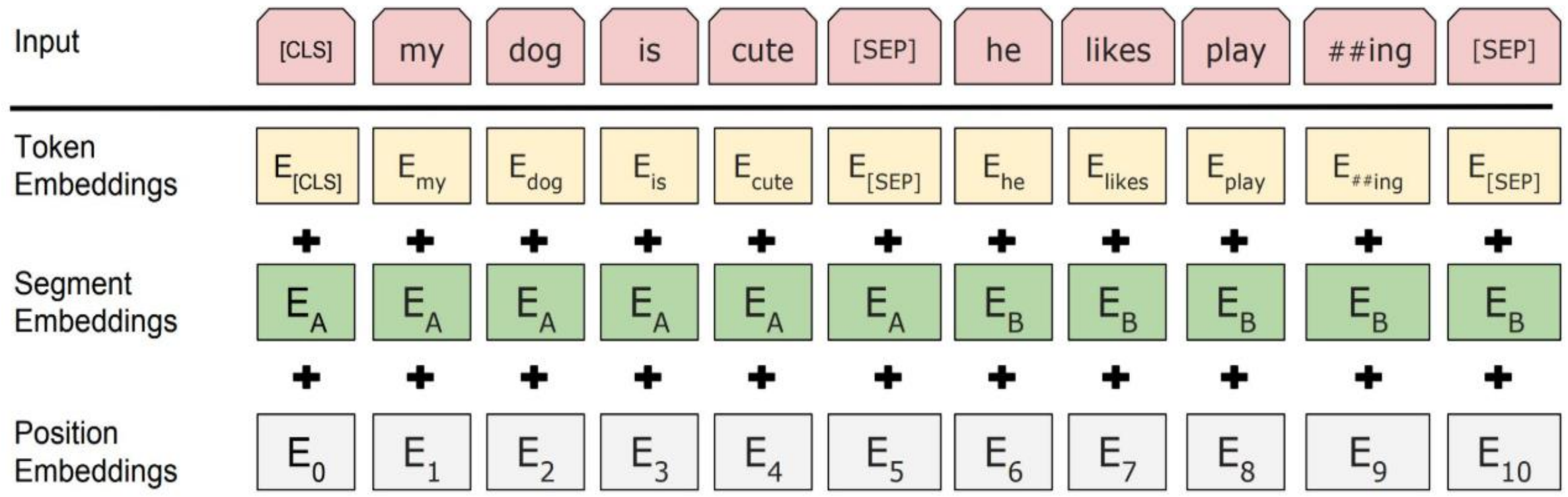

**Figure 22:** BERT Input Representation

The proposed BERT model implementation attempts to classify text into positive or negative sentiments to fine-tune BERT for SA. This is accomplished by using the Hugging Face Transformers library, which builds and trains models using TensorFlow.
Tokenizing text into input IDs and attention masks is the initial step. The data is truncated to a fixed length of 128 tokens for a uniform input size.
BERT's tokenizer encodes each sentence using special tokens such as [CLS] at the start and [SEP] at the end of each sentence, and padding ensures all sentences have the same length for efficient batch processing. For fine-tuning, a Keras model incorporates the pre-trained BERT model. The final classification is achieved by passing the output through a dense layer using a softmax activation function.
The BERT model uses input IDs and attention masks to create embeddings, using pooled output as a fixed-size tensor of shape (batch_size, 768), where 768 is the hidden side of the BERT base model
A dense layer with a softmax activation function is added to the BERT embeddings, mapping the 768-dimensional pooled output to the desired binary sentiment classification classes (positive and negative).
The BERT model compilation is done using the Adam optimizer with a 1e-5 learning rate. The loss function is categorical cross-entropy, suitable for multi-class classification, and the metric for evaluation is categorical accuracy, which measures the proportion of correctly classified samples.
The BERT model is trained across four epochs (with a batch size of 32), allowing the model to process 32 samples at a time before changing the weights. Training is done using tokenized training data, and validation is done using the 'fit' method.
After the training, the model's high categorical accuracy shows how proficient it is at proficiently classifying the sentiments.
BERT's ability to capture deep contextual representations and understand intricate relationships in the text makes it highly effective for SA. The BERT model for SA is successfully fine-tuned by this implementation. It can reliably determine sentiment even in more complex and nuanced text, exceeding many traditional methods and producing cutting-edge findings.

## 3.5 Presentation of the Result:

The presentation of results includes the prediction results. Different metrics are used to evaluate the performance of the proposed models. The experimental results are shown in section 4.

# 4. RESULT & DISCUSSION:

## 4.1. Performance Metrics

The proposed work's results are assessed using various statistical metrics listed below.
This confusion matrix (CM), a performance evaluation tool, shows the number of True Positives (TP), True Negatives (TN), False Positives (FP), and False Negatives (FN), indicating how accurate a classification model is [67].

**Table 4**: Confusion Matrix

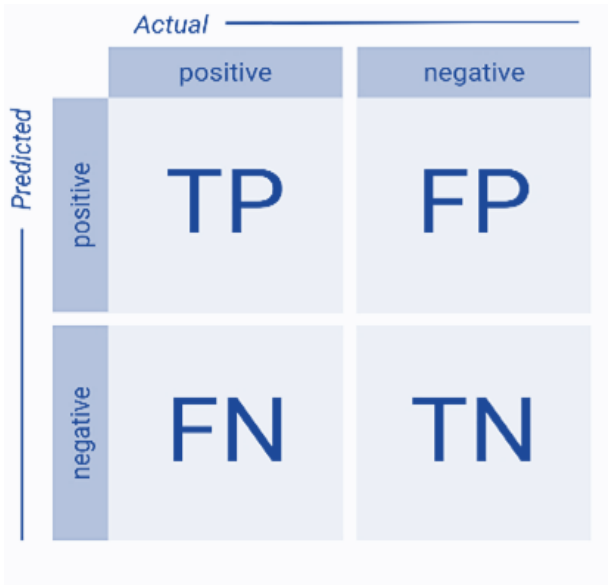

| Predicted \ Actual | positive | negative |
|---|---|---|
| positive | TP | FP |
| negative | FN | TN |

From Table 4, Accuracy, Precision, Recall, and F1-Score are defined in (Fig- 23).

**Figure 23**: Various Performance Metrics

$$\text{Accuracy} = (TP + TN)/\ (TP + TN + FN + FP)$$
$$\text{Precision} = TP/\ (TP + FP)$$
$$\text{Recall} = TP/\ (TP + FN)$$
$$\text{F1-Score} = 2 \times (\text{Recall} \times \text{Precision})/\ (\text{Recall} + \text{Precision})$$

Support is the quantity of actual occurrences for each class.
The evaluation includes generating predictions, generating learning curves, calculating a confusion matrix, comparison of different proposed models, and generating a classification report that provides detailed performance metrics.

## 4.2 Model Evaluation:

The DL models predict the sentiment using the ‘predict’ method. The predicted probabilities are then converted to a one-hot encoded format to match the format of the test labels. A classification report is produced, containing detailed performance metrics like accuracy, precision, recall, f1-score, and support for each class (negative and positive sentiment).

- *Baseline Model (NB) Evaluation:*

For data with a 90-10 split, the NB model as the baseline model obtained the maximum accuracy of **83**%. Table 5 shows the classification report of the NB model.

**Table 5: Classification Report for NB:**

| | ***Precision*** | ***Recall*** | ***F1-score*** | ***Support*** |
|---|---|---|---|---|
| *Negative* | 0.93 | 0.75 | 0.83 | 14659 |
| *Positive* | 0.74 | 0.92 | 0.82 | 10984 |
| ***Accuracy*** | | | **0.83** | 25643 |
| *Macro avg.* | 0.83 | 0.84 | 0.82 | 25643 |
| *Weighted avg.* | 0.85 | 0.83 | 0.83 | 25643 |

The performance of the NB algorithm is not so bad. For the classes (negative and positive emotions), the F1 score is around 83%. In particular, the overall accuracy is **83%**. The CM is shown in (Fig-25).

- *Advance DL Models Evaluation:*

**Table 6: Classification Report for CNN:**

| | *Precision* | *Recall* | *F1-score* | *Support* |
|---|---|---|---|---|
| *Negative* | 0.99 | 0.97 | 0.98 | 16225 |
| *Positive* | 0.96 | 0.98 | 0.97 | 11555 |
| ***Accuracy*** | | | 0.98 | 27780 |
| *Macro avg.* | 0.97 | 0.98 | 0.97 | 27780 |
| *Weighted avg.* | 0.98 | 0.98 | 0.98 | 27780 |

**Table 7: Classification Report for BiLSTM:**

| | *Precision* | *Recall* | *F1-score* | *Support* |
|---|---|---|---|---|
| *Negative* | 0.98 | 0.98 | 0.98 | 32710 |
| *Positive* | 0.97 | 0.97 | 0.97 | 22850 |
| ***Accuracy*** | | | 0.97 | 55560 |
| *Macro avg.* | 0.97 | 0.97 | 0.97 | 55560 |
| *Weighted avg.* | 0.97 | 0.97 | 0.97 | 55560 |

**Table 8: Classification Report for BERT:**

| | *Precision* | *Recall* | *F1-score* | *Support* |
|---|---|---|---|---|
| *Negative* | 1.00 | 0.99 | 0.99 | 14659 |
| *Positive* | 0.99 | 0.99 | 0.99 | 10984 |
| *Micro avg.* | 0.99 | 0.99 | 0.99 | 25643 |
| *Macro avg.* | 0.99 | 0.99 | 0.99 | 25643 |
| *Weighted avg.* | 0.99 | 0.99 | 0.99 | 25643 |
| *Sample avg.* | 0.99 | 0.99 | 0.99 | 25643 |

- *Accuracy for Train, Validation, and Test Sets:*

*72% for Train, 18% Validation and 10% for Testing*

| ***Metric*** | ***Table 9: CNN Result*** | | |
|---|---|---|---|
| | ***Train*** | ***Validation*** | ***Test*** |
| ***Accuracy*** | 98.94% | 99.28% | **97.56%** |

*70% for Train, 15% Validation and 15% for Testing*

| ***Metric*** | ***Table 10: BiLSTM Result*** | | |
|---|---|---|---|
| | ***Train*** | ***Validation*** | ***Test*** |
| ***Accuracy*** | 97.91% | 97.32% | **97.35%** |

The proposed CNN model obtained the maximum accuracy on the test set of 97.56%, whereas the proposed BiLSTM model obtained 97.35%.

*72% for Train, 18% Validation and 10% for Testing*

| ***Metric*** | ***Table 11: BERT Result*** | | |
|---|---|---|---|
| | ***Train*** | ***Validation*** | ***Test*** |
| ***Accuracy*** | 99.60% | 99.36% | **99.23%** |

The proposed BERT model obtained the maximum accuracy of 99.23%.

Advance DL models got higher accuracy than our proposed NB traditional ML model. We can see our proposed BERT model obtained the highest accuracy of **99.23%** amongst other proposed models.

So, we can definitely say that the advance BERT model is superior to other proposed models.

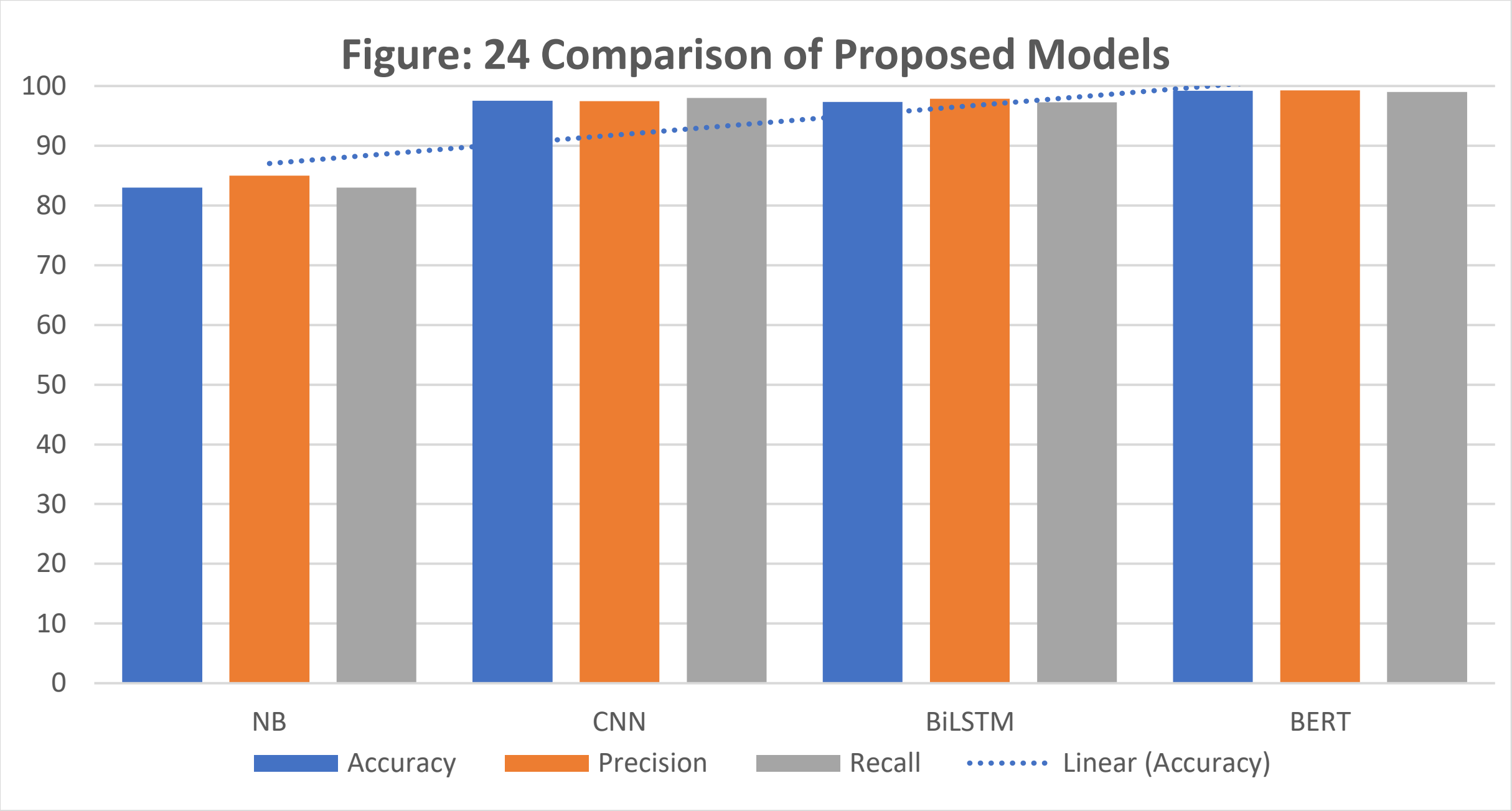

**Figure 24:** Comparison of proposed models

The BERT model has the highest accuracy of around 99%, which is shown in (Fig- 24).

- *Confusion Matrices:*

The confusion metrics for the proposed models with good accuracy are shown in (Fig- 25).

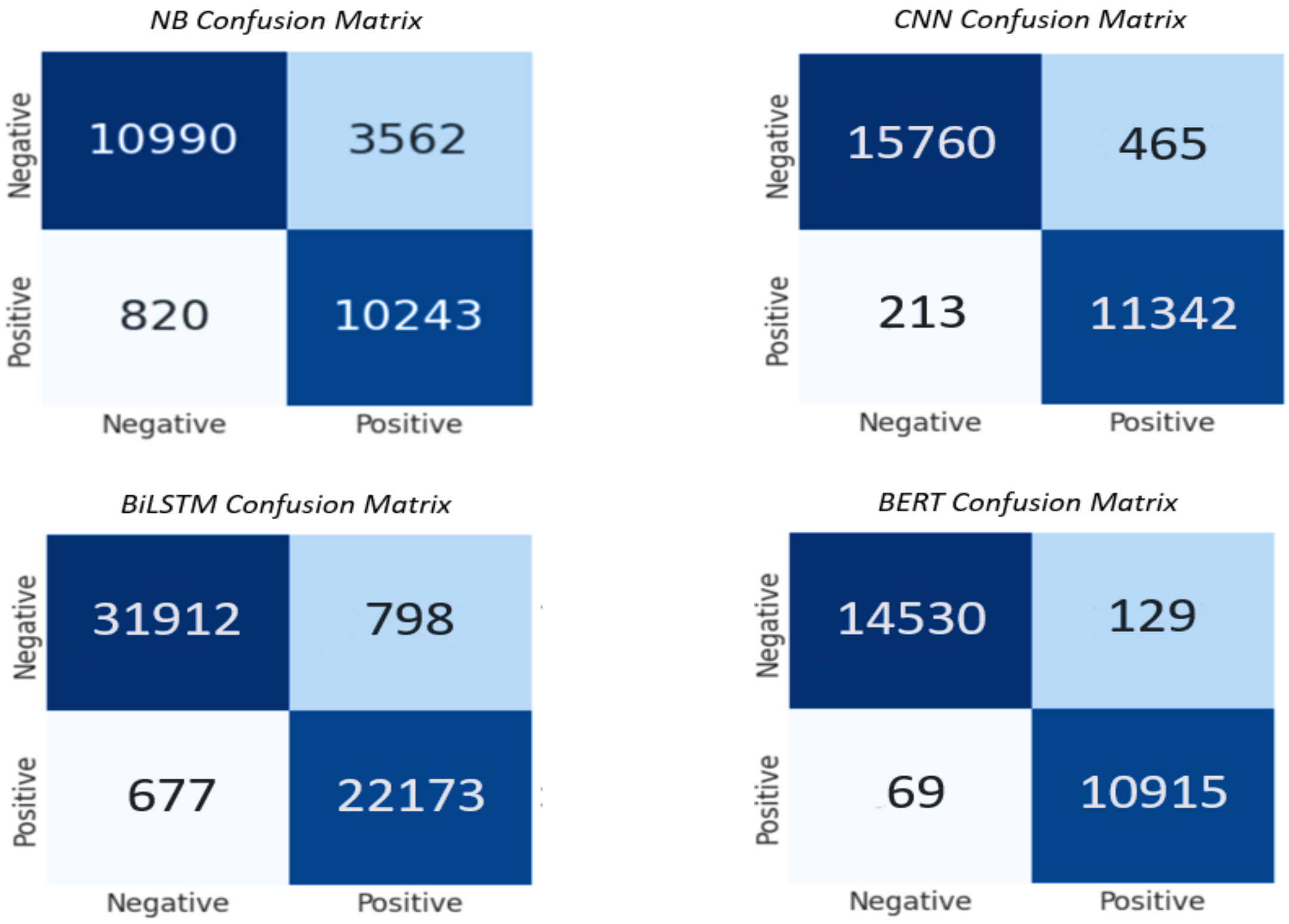

**Figure 25:** Confusion Matrices

- *Learning Curves:*
    - *Learning Curve for CNN Model:*

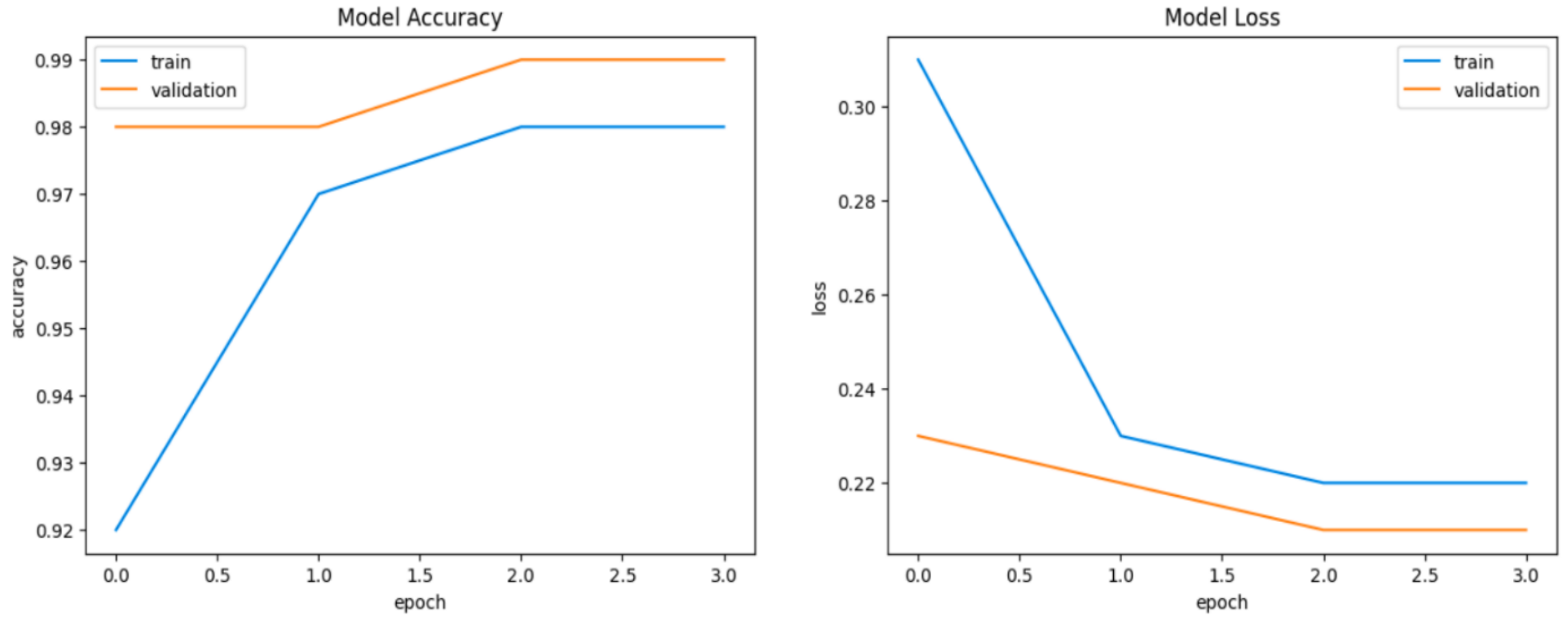

    - *Learning Curve for BiLSTM Model:*

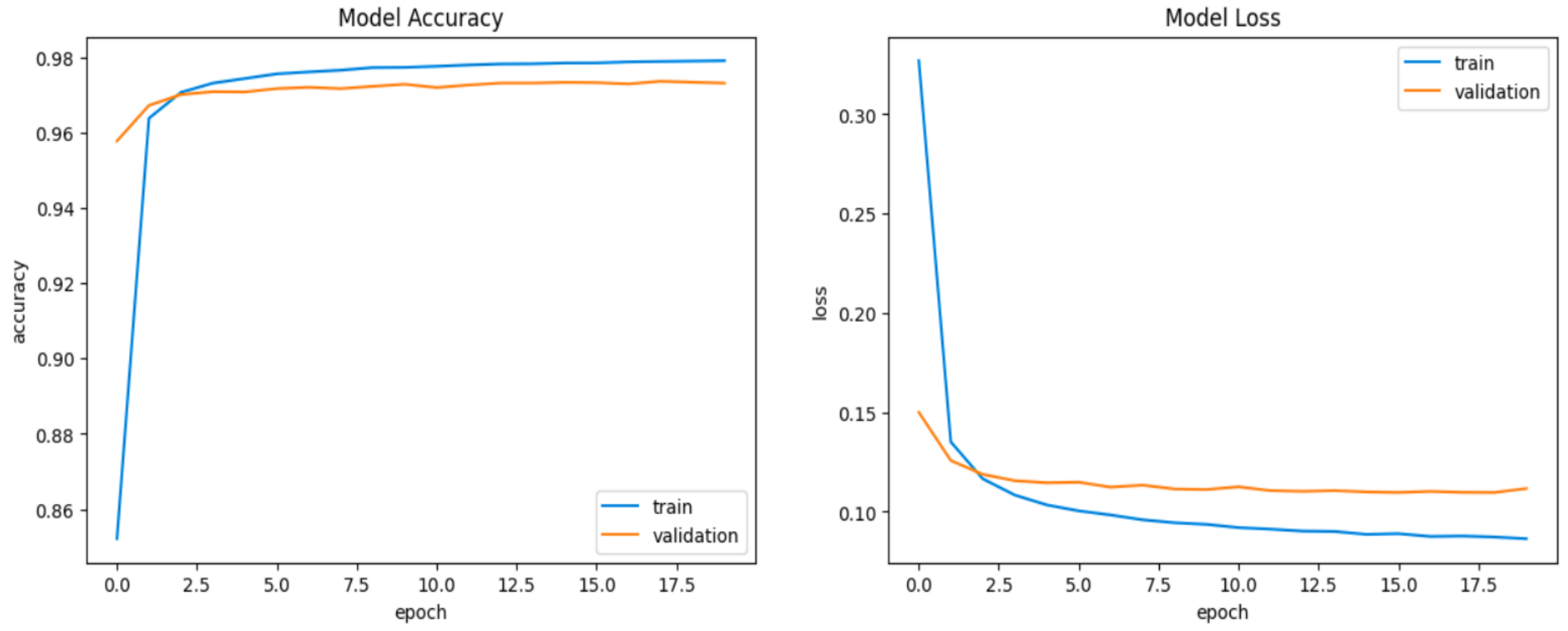

    - *Learning Curve for BERT Model:*

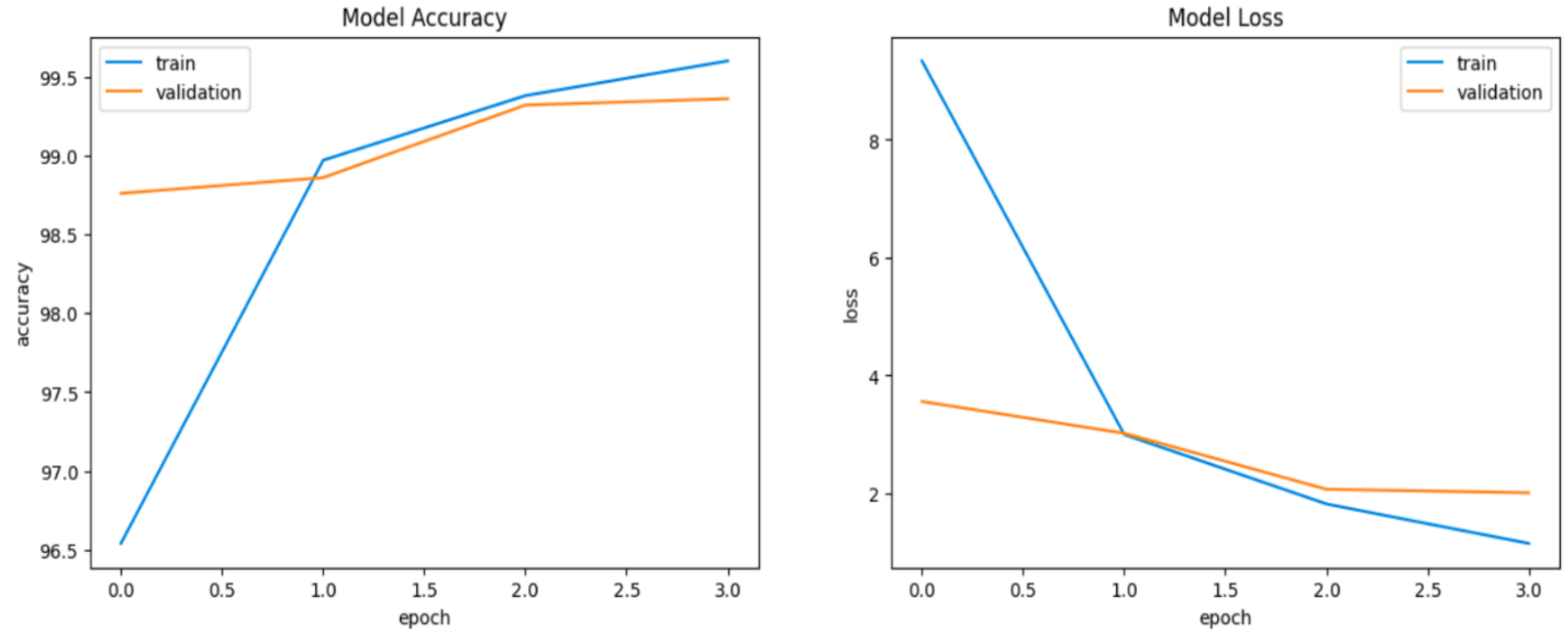

**Figure 26:** Learning Curves

The learning curves provide a better understanding of the dynamics of training and the DL model's performance. In our study, the BERT model outperformed CNN and BiLSTM, showing its superior ability to handle the complexities of SA in Olympic-related tweets. The learning curves underscored BERT's robust learning capabilities and highlighted the importance of monitoring these curves to make data-driven decisions for model optimization.

SA using DL models addresses some complexities of human language, but there are still challenges to generating reliable predictions. Below are scenarios where the SA method encounters difficulties:

1. A term with a positive or negative connotation may have an opposing orientation, e.g., "The old Olympic well is now dry and abandoned. The word "well" is positive, but the sentiment is negative. Emotional words might not always express true sentiment.
2. Sentiments can be expressed without explicitly sentimental words, e.g., "Watching the marathon in the pouring rain brought back bittersweet memories." Although lacking clear negative or positive words, the statement conveys a negative sentiment. Neutral sentences can also be interpreted as positive with low weightage.
3. This approach leaves out information from face-to-face interactions [68], such as tone of voice and facial expressions, which are crucial for understanding sentiments.

***"Sentiment words do not always express sentiments, as the context can change the orientation of the words, making it difficult for models to accurately predict the sentiment of a sentence" – Liu*** [69].

While our BERT model captures the context of sentences more effectively than basic approaches, these challenges highlight the intrinsic issues in sentiment expression and interpretation. Recognizing these limitations is essential for improving SA models further.

***"BERT has revolutionized sentiment analysis by capturing intricate relationships between words in context, enabling more nuanced understanding of language nuances and sentiment expression." -Jacob Delvin*** [70].

## 4.3 Comparative Analysis of Previous Work

Table 12 displays the comparative summary of the related work. Our comprehensive data preprocessing, robust training methodology and effective use of attention mechanisms collectively contribute to BERT's superior ability to capture and analyze complex and context-rich sentiments in Olympics tweets. These factors make BERT a highly effective tool for SA in this domain, resulting in superior accuracy compared to prior studies.

**Table 12:** Comparison with the previous works

| *Author and Year* | *Method* | *Dataset* | *Accuracy* |
|---|---|---|---|
| *Salau et al., 2023[22]* | LSTM | Olympics Twitter dataset | 88.20% |
| *Mello et al., 2022 [25]* | BERT | Olympics news articles dataset | 74.7% |
| *Alharbi and De Doncker, 2019 [32]* | LSTM,<br>CNN | SemEval-2016 Twitter dataset | 88.13%,<br>88.71%. |
| *PROPOSED METHOD* | BERT | Olympics Twitter dataset | **99.23%** |

# 5. CONCLUSION:

In conclusion, our work clearly shows that the advance DL models considerably improve SA accuracy. With a comprehensive diligent methodology and state-of-the-art models such as CNN, BiLSTM, and BERT, we achieved remarkable accuracy, the highest is 99.23% with the BERT model. This high accuracy shows how well BERT can capture complex, context-rich sentiments that are expressed in tweets. This SA study shows primarily positive emotions regarding the Olympics and highlights the significance of the event in promoting international unity, and global harmony and showing athletic excellence. Word clouds and frequency distributions are examples of data visualizations that highlight important sports, well-known athletes, bands, and many more of the spirit of the Olympics as a whole. Unigrams and bigrams give insight into current conversations and public opinion, expressing both enthusiasm and anxiety.
This analysis can help stakeholders such as event organizers, sponsors, and policymakers to understand public sentiment and make informed decisions to upgrade the Olympic experience. Future studies might look into other domains and other context-aware models to further refine SA capabilities. The information gathered from this study helps to develop deeper understandings and more sophisticated SA tools.

**Availability of data and material**

'All data generated or analyzed during this study will be available on request.'

**Competing interests**

'The authors declare that they have no known competing financial interests or personal relationships that could have appeared to influence the work reported in this paper.'

**Acknowledgments**

We are deeply grateful to the University of Burdwan's computer science department for their support in this research.